\documentclass{article}

\PassOptionsToPackage{numbers, compress}{natbib}

\usepackage[preprint]{neurips_2026}

\usepackage{multirow}
\usepackage{graphicx}
\usepackage{colortbl}
\usepackage{graphicx}
\usepackage{subcaption}
\usepackage{xspace}
\usepackage{multirow}
\usepackage{makecell}
\usepackage{algorithm}
\usepackage[normalem]{ulem}
\usepackage[export]{adjustbox}
\usepackage{algorithm}
\usepackage{algpseudocode}

\definecolor{yellow}{rgb}{1, 1, 0.7}
\definecolor{orange}{rgb}{1, 0.85, 0.7}
\definecolor{tablered}{rgb}{1, 0.7, 0.7}
\definecolor{red}{rgb}{1, 0, 0}

\definecolor{wincolor}{rgb}{0.85, 0.0, 0.0}

\definecolor{darkyellow}{rgb}{0.8, 0.8, 0.5}
\definecolor{darkred}{rgb}{0.7, 0.3, 0.3}
\definecolor{darkgreen}{rgb}{0.3, 0.7, 0.3}
\definecolor{blue}{rgb}{0.125, 0.469, 0.703}
\definecolor{lightskyblue}{rgb}{0.4706, 0.1922, 0.0000}
\definecolor{green}{rgb}{0, 1.0, 0}
\definecolor{pink}{rgb}{1, 0.4, 0.7}

\usepackage[utf8]{inputenc} 
\usepackage[T1]{fontenc}    
\usepackage{hyperref}       
\usepackage{url}            
\usepackage{booktabs}       
\usepackage{amsfonts}       
\usepackage{nicefrac}       
\usepackage{microtype}      
\usepackage{xcolor}         
\usepackage{graphicx}
\usepackage{wrapfig}

\title{NBS: No Bias Stereo}

\author{%
  Vage Taamazyan$^{*\,\dagger\,1}$ \quad
  Zhuowen Shen$^{*\,1,2}$ \quad
  Stefan Hinterstoisser$^{\ddagger\,1}$ \quad
  Alberto Dall'Olio$^{\ddagger\,1}$ \\[4pt]
  \bfseries                                  
  Agastya Kalra$^{1}$ \quad
  Aarrushi Shandilya$^{1}$ \quad
  Xin Li$^{2}$ \quad
  Wenping Wang$^{2}$ \quad
  Kartik Venkataraman$^{1}$ \\[6pt]
  \normalfont                                
  $^{1}$Intrinsic (Google) \qquad $^{2}$Texas A\&M University \\[4pt]
  {\normalfont\small
    $^{*}$co-first authors, order randomized \quad
    $^{\ddagger}$second authors \quad
    $^{\dagger}$project lead}
}

\begin{document}

\maketitle

\begin{figure}[H]
  \centering
  \includegraphics[width=\linewidth]{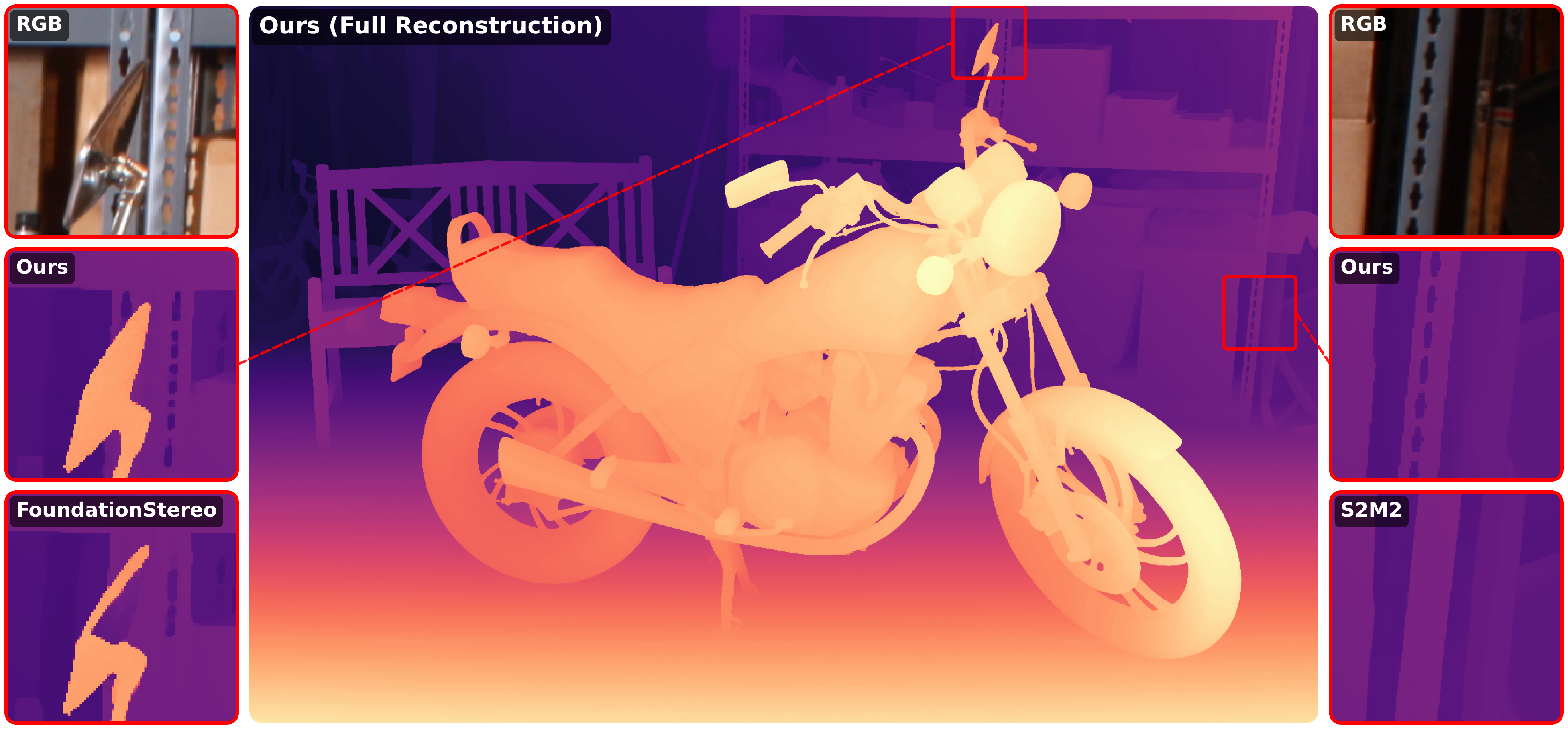}
  \caption{By removing complex inductive biases such as cost volumes and iterative GRU refinement, No Bias Stereo (NBS) accurately reconstructs fine details where prior methods fail.}
  \label{fig:teaser}
\end{figure}
\begin{abstract}
    Stereo reconstruction is one of the last remaining Computer Vision tasks where all state-of-the-art methods employ a heavy architectural inductive bias. Even though it has been demonstrated that the task can be solved using general-purpose methods, it is widely believed that inductive biases in stereo are strictly necessary for both high-quality results and computational efficiency. We challenge this paradigm. In this paper, we demonstrate that both state-of-the-art accuracy and superior runtime efficiency are achievable with a model completely devoid of architectural inductive biases, relying instead on a simple, end-to-end Vision Transformer. By training on massive synthetic datasets, we show that pure data-driven learning can surpass explicitly engineered geometry. This work proves that explicit inductive biases are no longer a prerequisite for stereo matching, ultimately unlocking true scaling laws for continuous improvement in 3D reconstruction. Our project page is available at
\href{http://intrinsic.ai/publications/NBS}{intrinsic.ai/publications/NBS}.
\end{abstract} 

\section{Introduction}

Stereo reconstruction is a foundational computer vision task, one that has been significantly advanced by deep learning over the last decade. During this period, the community has seen a steady progression toward higher precision, driven by the introduction of increasingly sophisticated architectural components. Modern state-of-the-art models~\cite{wen2025foundationstereo, min2025s2m2, wang2024selective, min2025depthfocus, li2022practical} now rely on a highly specialized pipeline: deep feature extraction is followed by the construction of explicit 3D or 4D cost volumes, which are then processed via 3D convolutions or filtered through iterative update blocks—most notably Gated Recurrent Units (GRUs)—to produce high-resolution disparity maps. Consequently, squeezing out marginal improvements in 3D reconstruction has increasingly demanded ever more complex, custom-engineered modules.

Recently, this complexity has increased further as researchers have begun integrating "internet-scale" knowledge by fusing frozen, pre-trained monocular depth models into these already intricate stereo pipelines~\cite{wen2025foundationstereo, cheng2025monster, jiang2025defom, zhou2025all}. Currently, all methods in the top 30 of the ETH3D~\cite{eth3d_stereo, schoeps2017cvpr} or Middlebury~\cite{scharstein2002taxonomy, middlebury_stereo, scharstein2014high} benchmarks rely on strong architectural inductive biases. As a result, the stereo vision community has somewhat diverged from the broader machine learning landscape, anchoring itself to highly specialized operations rather than embracing general-purpose architectures.

In this paper, we aim to bridge this gap. A central lesson from the broader deep learning community is that a simple, general-purpose architecture, when combined with sufficient data and compute, ultimately surpasses highly specialized models. We demonstrate that stereo reconstruction is no exception to this rule. Concurrently, the field has recognized that integrating stereo matching with strong monocular depth priors can significantly resolve ambiguities in occluded or textureless regions. We propose that a single, end-to-end Vision Transformer (ViT)~\cite{dosovitskiy2020image} is the optimal architecture to unify these two objectives. By eschewing rigid cost volumes and scaling our model on massive synthetic corpora, our transformer dynamically attends to both inter-image matching (stereo cues) and intra-image semantics (monocular priors). This pure attention mechanism naturally achieves the "best of both worlds" without requiring articulated sub-modules or explicit geometric filtering.

As a result, our model currently ranks first on the ETH3D benchmark~\cite{eth3d_stereo} and demonstrates state-of-the-art, highly competitive performance across several other benchmarks analyzed in this paper. Furthermore, by dispensing with computationally heavy cost volumes and iterative refinement steps, our architecture achieves highly efficient inference—operating several times faster than existing top-performing methods at both 1K (e.g., $960 \times 540$) and 2K (e.g., $1920 \times 1080$) resolutions while simultaneously delivering superior accuracy. A notable limitation, however, is scaling beyond these dimensions (e.g., to 4K resolution), as training at such extreme scales remains computationally demanding for the standard self-attention mechanism.

Ultimately, we hope this work elevates stereo reconstruction to the same paradigm as foundation models like Depth Anything~\cite{lin2025depth} and VGGT~\cite{wang2025vggt}, motivating the community to move beyond custom inductive biases and embrace true scaling laws. To this end, our main contributions are as follows:
\begin{itemize}
    \item \textbf{An Inductive-Bias-Free Foundation Model:} We present the first competitive stereo matching model devoid of explicit cost volumes, 3D convolutions, or iterative refinement, relying instead on a unified, end-to-end Vision Transformer architecture.
    \item \textbf{Emergent Monocular-Stereo Synthesis:} We demonstrate that our model internally learns to fuse geometric matching with monocular priors, achieving robust zero-shot generalization across diverse benchmarks and maintaining plausible depth estimation even under total camera occlusion.
    \item \textbf{Comprehensive Evaluations:} We evaluate our approach on the standard low-resolution ETH3D benchmark~\cite{eth3d_stereo}. However, due to the limited test set of ETH3D, we also evaluate both our model and prior methods on additional high-fidelity synthetic data generated via SimpleProc~\cite{ma2026fullyproceduralsyntheticdata}, as well as the real-world XYZ-IBD~\cite{huang2025xyzibdhighprecisionbinpickingdataset} industrial part dataset post-processed for stereo. These evaluations demonstrate that data-driven learning can surpass explicitly engineered geometric biases. To support reproducibility, we attach our SimpleProc scene-generation seeds and post-processed, stereo-rectified XYZ-IBD pairs with ground-truth disparity in the supplement.
\end{itemize}

\section{Related Work}

\subsection{Stereo}

Classical stereo matching algorithms relied on local patch matching or semi-global optimization like SGM~\cite{hirschmuller2008stereo}. Deep learning introduced a paradigm shift, replacing hand-crafted features with dense correlation volumes in architectures like GC-Net~\cite{kendall2017end} and DeepPruner~\cite{duggal2019deeppruner}. To regularize these volumes, extensive research established dominant 3D CNN filtering architectures, including PSMNet~\cite{chang2018pyramid}, AANet~\cite{xu2020aanet}, CFNet~\cite{shen2021cfnet}, and PCW-Net~\cite{shen2022pcw}. While accurate, explicitly building and filtering 4D cost volumes is notoriously compute- and memory-intensive. To mitigate this bottleneck for real-time, high-resolution processing, lightweight alternatives like CoEx~\cite{bangunharcana2021correlate}, fastACV~\cite{xu2023accurate}, and BGNet+~\cite{xu2021bilateral} were proposed, though often at the expense of high-resolution precision.

To bypass heavy 3D convolutions, RAFT-Stereo~\cite{lipson2021raft} introduced a highly successful alternative: GRU-based iterative refinement over a 2D correlation field. This optical-flow-inspired approach was rapidly enhanced. CREStereo~\cite{li2022practical} proposed cascaded refinement for real-world distortions, while hybrid architectures like IGEV-Stereo~\cite{xu2023iterative} and IGEV++~\cite{xu2024igev++} integrated geometry encoding volumes to guide disparity propagation. Subsequent works further refined this recurrent framework via dynamic frequency selection~\cite{wang2024selective}, motif-based attention~\cite{chen2024mocha}, and continuous disparity updates~\cite{tosi2024neural}. Despite these advances, the localized nature of iterative search limits the ability to resolve global ambiguities. 

To overcome this limited receptive field, the field has increasingly integrated Vision Transformers~\cite{dosovitskiy2020image} and large-scale pretraining. Early global matching architectures like STTR~\cite{li2021revisiting}, CSTR~\cite{guo2022context}, and UniMatch~\cite{xu2023unifying} replaced explicit cost volumes with self- and cross-attention mechanisms. Concurrently, methods utilizing cross-view completion pretraining~\cite{weinzaepfel2023croco, kim2025unitt} and full-context transformers~\cite{bengana2022seeking, jia2024transformer} proved highly effective. To explicitly bridge the sim-to-real gap, recent frameworks~\cite{wen2025foundationstereo, wen2025fastfoundationstereo, bartolomei2024stereo, min2025s2m2, wang2026waft} inject internet-scale monocular depth priors. However, despite leveraging global attention and massive priors, these hybrid approaches often retain specialized heuristics—such as multi-view matching modules or explicit cost filtering—keeping them fundamentally tied to the inductive biases of localized stereo geometry.

\subsection{Foundation Models}

The broader vision community has embraced generalized foundation models that tackle diverse tasks using unified architectures. Initiated by the Vision Transformer (ViT)~\cite{dosovitskiy2020image}, this trajectory extended to dense spatial prediction via DPT~\cite{ranftl2021dpt} and self-supervised models like DINOv2~\cite{oquab2024dinov} and DINOv3~\cite{simeoni2025dinov3}, which provide robust, task-agnostic features. Capitalizing on these representations, recent monocular geometry estimation frameworks~\cite{kang2024depthanythingv2, hu2024metric3dv2, bochkovskiy2024depthpro, wang2024moge, wang2025moge2} have achieved unprecedented accuracy, effectively laying the groundwork for unified multi-view geometric reasoning.

Leveraging these foundational representations, recent advancements focus increasingly on multi-view dense prediction. Feed-forward models like DUSt3R~\cite{wang2024dust3r} and its highly scalable successor Fast3R~\cite{yang2025fast3r} have abandoned traditional explicit multi-view constraints for direct, unprojected pointmap regression. Expanding on this paradigm, VGGT~\cite{wang2025vggt} grounds transformer outputs in dense visual geometry, while Depth Anything 3~\cite{lin2025depth} obviates architectural specialization entirely, using a vanilla DINO transformer to predict unified depth-ray representations from arbitrary views. Our approach aligns closely with this minimalist, data-driven trajectory. By dispensing with bespoke cost-volume engineering and scaling a foundational ViT on massive synthetic corpora, we underscore the latent capability of simple feed-forward transformers in stereo reconstruction. Crucially, while embracing this general-purpose philosophy, we demonstrate that our model achieves significantly higher precision in stereo depth estimation than these broader 3D foundation models.

\section{Method}
\label{sec:method}

\begin{figure*}[t]
  \centering
  \includegraphics[width=\linewidth]{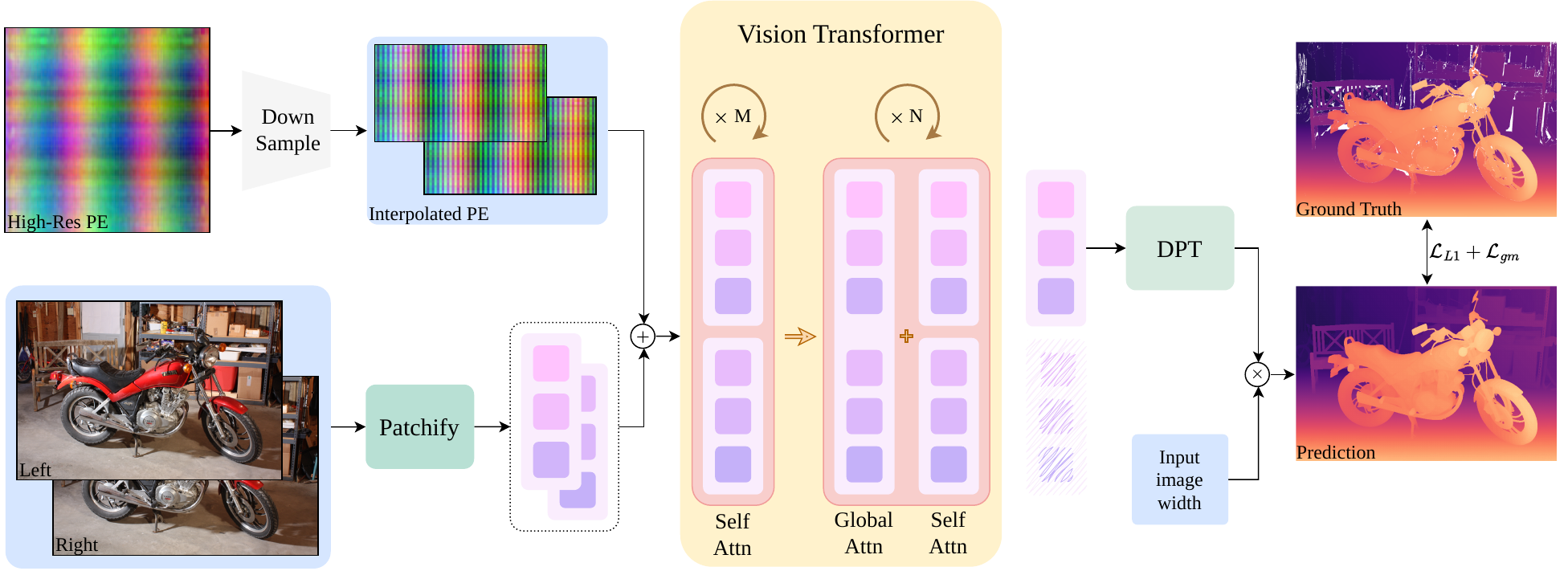}
  \caption{Our proposed architecture maintains a streamlined design. It leverages a standard Vision Transformer backbone -- integrating global and local self-attention mechanisms -- alongside a DPT decoder. Previous state-of-the-art stereo methods~\cite{wen2025foundationstereo, min2025s2m2} use complex inductive biases such as frozen foundation models, multi-resolution cost volumes and iterative refinement via GRUs.}
  \label{fig:model}
\end{figure*}

In this section, we discuss the model architecture illustrated in Figure~\ref{fig:model}, detailing our approach to high-resolution inference, the training data composition, and the overall training pipeline.

\subsection{Task}
We solve a standard stereo reconstruction task: given a pair of rectified images, we reconstruct a disparity map, which represents the horizontal pixel displacements between the left and right images. We assume that the disparity values are strictly positive and that disparities are well-defined even for regions in the left image that lack a direct match in the right image, as disparity is inherently a function of depth.

\subsection{Model Architecture and High-Resolution Inference}
\label{sec:model}
Building upon DINOv2~\cite{oquab2024dinov} with registers, we employ a standard Vision Transformer (ViT-Large)~\cite{dosovitskiy2020image} encoder, featuring both local and global self-attention layers similar to the designs described in VGGT~\cite{wang2025vggt} and Depth Anything 3~\cite{lin2025depth}. The combination of local and global self-attention allows the network to dynamically balance cross-image matching and intra-image semantics. This enables reliable disparity estimation across all parts of the image, including occluded regions or areas with no reliable match, such as transparent, reflective, or textureless surfaces. Specifically, our architecture employs local self-attention for layers 1 through 7, while every alternating layer from 8 to 24 is designated for global attention. For the initialization of the encoder weights, we leverage the original pretrained weights from DINOv2~\cite{oquab2024dinov}.

The task of disparity estimation requires very fine-grained positional understanding. To accommodate this, we find it highly beneficial to leverage positional embeddings with higher spatial resolution than the original DINOv2 model. Specifically, we use a $148 \times 148$ spatial resolution for $966 \times 546$ image resolution and $296 \times 296$ spatial resolution for $1932 \times 1330$ image resolution. Similar to DINOv2, the positional embeddings are resampled to the tokenized image's spatial resolution during the forward pass. We initialize the positional embeddings by upsampling the $37 \times 37$ DINOv2 pretrained embeddings.

To predict the disparity map, we utilize a Dense Prediction Transformer (DPT~\cite{ranftl2021dpt}) decoder on the embeddings of the left input image. The DPT decoder effectively aggregates dense, multi-scale feature maps from the encoder to output fine-grained, edge-aware disparity predictions. The network is designed to output normalized disparity values between 0 and 1, which are subsequently multiplied by the image width to obtain the absolute disparity values in pixels. Forcing the model to generate normalized outputs relative to the image dimensions ensures that the learned disparity representations remain inherently independent of the specific input resolution.

The integration of a subsequent fine-tuning stage for specific, fixed higher resolutions (as explained in Section~\ref{sec:high_res}), in conjunction with the aforementioned architectural choices, enables the model to seamlessly process images resized to the model's maximum native resolution of $1932 \times 1330$ during inference.

\subsection{Internal Training Data}
To help satisfy the substantial data requirements inherent to Vision Transformers, we augmented our training mixture with a large-scale internal synthetic stereo dataset consisting of approximately 2.4 million scenes. This dataset features stereo pairs with varying baselines, randomized camera intrinsics, diverse 3D assets, varied lighting conditions, material properties, and random object placements. Rendering was performed at three standard resolutions: $1024 \times 768$, $1280 \times 720$, and $1920 \times 1080$. Unlike the open-source stereo datasets utilized in our pipeline, this internal corpus introduces diversity in both resolution and aspect ratio. Although this proprietary dataset will not be publicly released—and is therefore not claimed as a contribution of this work—it serves as a valuable supplement to our data mixture. Ultimately, we find that combining this internal data with diverse public datasets helps provide the sheer scale and dimensional variety necessary to effectively optimize our ViT model.

\subsection{Training Process}
\label{subsec:training_process}
We employ a multi-stage training pipeline designed to progressively build strong geometric priors, adapt to high resolutions, and fine-tune for specific real-world benchmarks.

\subsubsection{Base Model Pretraining}
\label{sec:base_model}
During the initial phase, we initialize the ViT-Large backbone with DINOv2~\cite{oquab2024dinov} weights and pretrain our base model at a fixed resolution of $966 \times 546$, utilizing an upsampled positional embedding grid of $148 \times 148$. This stage is trained for 170,000 steps exclusively on the FSD~\cite{wen2025foundationstereo} dataset. Training is distributed across 128 A100 GPUs using mixed precision (bf16). We optimize the network using the Muon-AdamW~\cite{liu2025muonscalablellmtraining, loshchilov2019decoupledweightdecayregularization} optimizer with a peak learning rate of 1.46e-4, applying a 7,000-step linear warmup followed by a cosine annealing schedule.

\subsubsection{High-Resolution Adaptation}
\label{sec:high_res}
To enable robust high-resolution inference, we branch from the base model in~\ref{sec:base_model} and employ a multi-step pipeline that gradually scales the training resolution to a final target of $1932 \times 1330$. To accommodate the expanded token grid at this maximum resolution, the positional embeddings are progressively upsampled to a final spatial resolution of $296 \times 296$. Throughout this adaptation phase, the model is exposed to our massive, diverse mixture of 16 datasets, comprising our 2.4M scene internal dataset alongside 13 public datasets~\cite{wen2025foundationstereo, tremblay2018fallingthingssyntheticdataset, bao2020instereo2k, jing2024match, mehl2023spring, tosi2021smd, li2022practical, MIFDB16, cabon2020virtualkitti2, 10.1109/IROS45743.2020.9341801, patel2025tartanground, 9428423, yan2025proceduraldatasetgenerationzeroshot}. To preserve the rich semantic features established during pretraining, we apply a differential learning rate strategy, scaling the ViT encoder's learning rate down to 10\% of the overall rate. The intermediate scaling steps are trained using mixed precision (bf16) on 128 A100 GPUs, while the final high-resolution polish is conducted in full precision (float32) on 64 H100 GPUs to ensure peak predictive fidelity.

\subsubsection{Benchmark-Specific Fine-Tuning}
To achieve state-of-the-art results on specific real-world benchmarks, we take the pre-trained base model ($966 \times 546$) in~\ref{sec:base_model} and perform a brief fine-tuning phase (36,000 steps) with a significantly reduced learning rate. For our ETH3D benchmark~\cite{eth3d_stereo} submission, this fine-tuning combines the FSD dataset with the official training sets of ETH3D~\cite{schoeps2017cvpr} and Middlebury~\cite{scharstein2014high}.

\subsubsection{Training Objective}

To optimize for global value accuracy, we employ a standard masked L1 loss between ground truth $\hat{d}$ and the prediction $d$, defined as:
\begin{equation}
    \mathcal{L}_{L1} = \frac{1}{\sum_{i} M_{i}} \sum_{i} M_{i} |\hat{d}_{i} - d_{i}|, \quad M_{i} = \mathbb{I}(d_i > 0).
\end{equation}
To encourage smoother gradient changes and sharper disparity discontinuities in the prediction, we incorporate a multi-scale gradient matching term $\mathcal{L}_{gm}$~\cite{li2018megadepth, ranftl2020towards}. This is defined as an $\ell_1$ penalty on the spatial gradients of the disparity difference map between the prediction and the ground truth:
\begin{equation}
    \mathcal{L}_{gm} = \frac{1}{\sum_{i} M_{i}} \sum_{i} \sum_{k} M_{i} (|\nabla_x D_i^k| + |\nabla_y D_i^k|),
\end{equation}
where $D_i^k$ is the disparity difference map at position $i$ and scale $k$. The final loss is the sum of these two terms.

\section{Results and Experiments}
\label{sec:results}

\subsection{Benchmark Evaluation}

We evaluate our model on the standard low resolution stereo reconstruction benchmark: ETH3D~\cite{eth3d_stereo}. However, because the benchmark is nearing saturation and contains a limited number of scenes, we extend our evaluation to two novel benchmarks designed to thoroughly test model generalizability. The first is a purely synthetic dataset generated via SimpleProc~\cite{ma2026fullyproceduralsyntheticdata}. Because this recently released procedural tool was not used to train any current state-of-the-art models (including our own), it serves as a strict "out-of-distribution" benchmark. Our second additional benchmark is derived from the XYZ-IBD~\cite{huang2025xyzibdhighprecisionbinpickingdataset} dataset, repurposed specifically for stereo evaluation. To ensure full reproducibility, we include the processed XYZ-IBD dataset and the SimpleProc generation seeds in the supplement.

\begin{figure}[H]
  \centering
  \includegraphics[width=\linewidth]{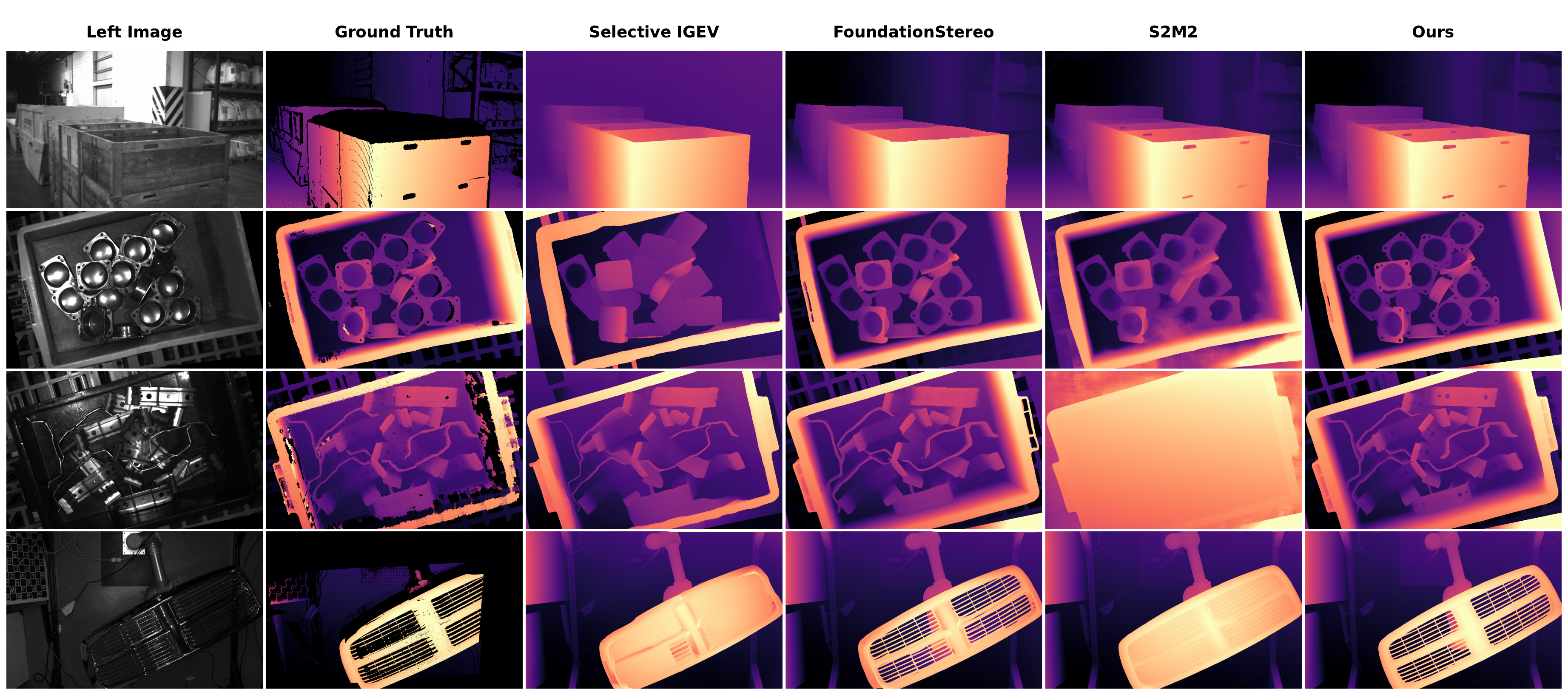}
  \caption{Our method achieves superior visual quality and state-of-the-art results, successfully reconstructing fine details that others miss, such as the holes in the containers (row 1), the complex shapes of cluttered objects (rows 2 and 3), and the shape and thin structures of the grill (row 4).}
  \label{fig:comparison}
\end{figure}

We employ standard industry metrics across all evaluations: End-Point Error (EPE), Root Mean Square error (RMS), and the bad@$X$ metric (where $X \in \{0.5, 1, 2, 4\}$ denotes the pixel error threshold). Comparative results on the benchmarks are summarized in Table~\ref{tab:cross_dataset_evaluation}, and qualitative reconstruction examples are presented in Figure~\ref{fig:comparison}.

\subsubsection{SimpleProc Evaluation Dataset}
\label{subsec:simpleproc_mainpaper}
We utilized SimpleProc~\cite{ma2026fullyproceduralsyntheticdata} to generate a procedural dataset at two resolutions: SimpleProc-S with 661 scenes at $966 \times 546$ pixels, and SimpleProc-M with 530 scenes at $1932 \times 1092$ pixels. To tailor this tool for stereo at desired resolutions, we modified the default scene generation pipeline as explained in the supplement. 
To prevent selection bias, all scenes were generated in a single batch without any manual filtering. While the randomized geometric primitives in these scenes do not test high-level semantic understanding, they provide a rigorous, unbiased assessment of a model's fundamental matching capabilities across a wide variety of unconstrained shapes and materials. 

\subsubsection{Industrial Parts Evaluation Dataset}
\label{subsec:xyz_mainpaper}
Our second additional benchmark is constructed from XYZ-IBD~\cite{huang2025xyzibdhighprecisionbinpickingdataset}, an open-source dataset of industrial parts captured via a structured-light 3D scanner and a multi-camera setup. Although designed for 6-DoF pose estimation, its high-fidelity captures of singulated and heavily cluttered parts make it ideal for stereo evaluation. From the five images available per scene, we select and rectify the two that most closely resemble a left-right stereo configuration. We then project all five structured-light scans onto the left image, fusing them into a ground-truth depth map via a per-pixel median operation. After manually removing scenes with large artifacts, we establish a final evaluation set of 55 annotated stereo pairs, which we provide in the supplement.

\subsubsection{Quantitative Results}
As demonstrated in Table~\ref{tab:cross_dataset_evaluation}, our model achieves state-of-the-art performance across all four evaluation datasets. Whether processing the established real-world scenes of ETH3D, the complex reflective materials of XYZ-IBD, or the unconstrained procedural geometries of SimpleProc, our inductive-bias-free architecture consistently exhibits superior matching precision. On the SimpleProc benchmark, for example, our model demonstrates significantly greater overall robustness as the error margin widens. Specifically, it outperforms all baselines across thresholds from bad@1.0 to bad@4.0, reducing the severe error rate of the next best model, S2M2~\cite{min2025s2m2}, by nearly 50\%. This robust, high-fidelity disparity estimation generalizes across both real and synthetic domains, validating the efficacy of our general-purpose attention mechanism over explicitly engineered geometry.

\begin{table}[htbp]
\centering
\caption{Quantitative cross-dataset evaluation. We report EPE, bad@1.0, and bad@4.0 across several benchmarks. Our model achieves the best performance in the majority of categories.}
\label{tab:cross_dataset_evaluation}
\resizebox{\textwidth}{!}{%
\begin{tabular}{l ccc ccc ccc ccc}
\toprule
\multirow{2}{*}{Model} & \multicolumn{3}{c}{ETH3D~\cite{eth3d_stereo}} & \multicolumn{3}{c}{SimpleProc-S~\cite{ma2026fullyproceduralsyntheticdata}} & \multicolumn{3}{c}{SimpleProc-M~\cite{ma2026fullyproceduralsyntheticdata}} & \multicolumn{3}{c}{XYZ-IBD} \\ 
\cmidrule(lr){2-4} \cmidrule(lr){5-7} \cmidrule(lr){8-10} \cmidrule(lr){11-13}
 & EPE & b@1.0 & b@4.0 & EPE & b@1.0 & b@4.0 & EPE & b@1.0 & b@4.0 & EPE & b@1.0 & b@4.0 \\ 
\midrule
CREStereo~\cite{li2022practical}   & 0.14 & 1.09 & 0.12 & 0.33 & 4.72 & 1.43 & 0.64 & 8.03 & 2.75 & 15.92 & 62.28 & 36.04 \\ 
CroCo~\cite{weinzaepfel2023croco}   & 0.15 & 1.14 & 0.18 & 0.45 & 6.94 & 1.82 & 0.91 & 9.58 & 3.75 & 34.98 & 84.94 & 67.95 \\ 
Sel-IGEV~\cite{wang2024selective}  & 0.15 & 1.56 & 0.28 & 0.52 & 6.13 & 2.07 & 0.75 & 8.16 & 3.19 & 17.48 & 60.92 & 34.98 \\ 
FS~\cite{wen2025foundationstereo}  & 0.13 & 0.48 & 0.21 & 0.48 & 4.25 & 1.95 & 0.61 & 6.27 & 3.22 & 15.75 & \textbf{54.69} & 28.35 \\ 
S2M2~\cite{min2025s2m2}            & 0.10 & 0.26 & 0.04 & 0.29 & 4.16 & 1.12 & 0.44 & 6.04 & 2.31 & 19.24 & 57.22 & 30.69 \\ 
\midrule
\textbf{Ours} & \textbf{0.09} & \textbf{0.16} & \textbf{0.02} & \textbf{0.25} & \textbf{3.03} & \textbf{0.63} & \textbf{0.40} & \textbf{4.22} & \textbf{1.26} & \textbf{11.39} & 55.61 & \textbf{23.39} \\
\bottomrule
\end{tabular}%
}
\end{table}

\begin{figure}[h]
  \centering
  
  \centering
  \begin{minipage}{0.56\textwidth}
    \centering
    \captionof{table}{Due to our simple transformer architecture being highly optimized on modern GPUs, our approach achieves 4x faster runtime and 2.8x lower peak memory at float-16 on A100s compared to the previous best methods (FS~\cite{wen2025foundationstereo}, S2M2~\cite{min2025s2m2}) on SimpleProc-S~\cite{ma2026fullyproceduralsyntheticdata}.}
    \resizebox{\textwidth}{!}{%
      
\begin{tabular}{@{}l ccccc}
    \toprule
Method           & Params (M) & TFLOPs & Runtime (s) & Mem (G) & bad@1       \\
    \midrule
CREStereo~\cite{li2022practical}
& \textbf{9.5}        & 9.5   & 0.693       & 2.23    & 4.72 \\
CroCo~\cite{weinzaepfel2023croco}
& 436.2      & 75.5   & 3.778       & 3.88    & 6.94 \\
\textbf{Ours}             & 351.5      & \textbf{3.7}    & 0.323       & 2.52    & \textbf{3.03} \\
\midrule
Sel-IGEV (F16)~\cite{wang2024selective}
& 13.1       & 18.2   & 0.662       & 1.35    & 6.13 \\ 
FS (F16)~\cite{wen2025foundationstereo}
& 374.5      & 25.9   & 0.872       & 6.74    & 4.25 \\
S2M2 (F16)~\cite{min2025s2m2}
& 405.7      & 11.3   & 0.240       & 3.52    & 4.16 \\
\textbf{Ours (F16)}        & 351.5      & \textbf{3.7}    & \textbf{0.060}      & \textbf{1.23}    & 3.24 \\
    \bottomrule
\end{tabular}
    }
    \label{tab:efficiency}
  \end{minipage}
  \hfill
  \begin{minipage}{0.4\textwidth}
    \centering
    \includegraphics[width=0.95\textwidth]{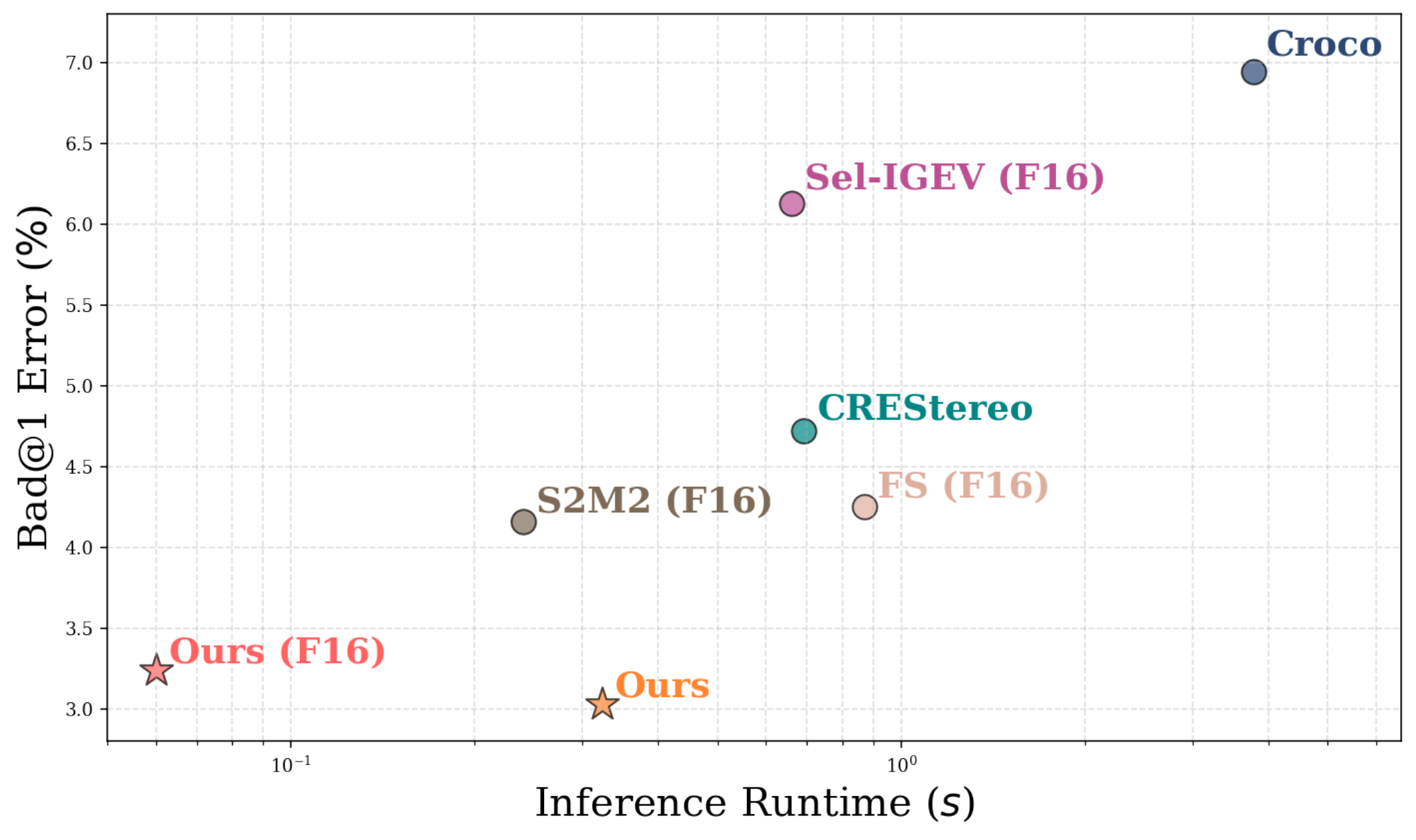}
    \caption{At $966 \times 546$ resolution, our simple architecture achieves superior runtime compared to previous state-of-the-art models.}
    \label{fig:efficiency}
  \end{minipage}

\end{figure}

Table~\ref{tab:efficiency} and Figure~\ref{fig:efficiency} present a comprehensive evaluation of model efficiency. All benchmarks were conducted on a single NVIDIA A100 GPU using float-32 precision, except where otherwise noted. Our proposed method demonstrates superior performance, achieving SoTA results while significantly reducing computational overhead, inference latency, and memory footprint. Furthermore, by leveraging Flash-Attention~\cite{dao2023flashattention} and float-16 (F16) precision, we achieve a $5.4\times$ acceleration in inference speed and a $2\times$ reduction in memory usage, with only a marginal impact on accuracy.

\subsection{Emergent Monocular Depth Capabilities}
We investigate the model's internal priors by evaluating its performance in a monocular depth estimation scenario. In this experiment, we provide a standard left image but replace the right image with a zero-initialized tensor, effectively simulating a total right-camera occlusion. We benchmarked our model against leading monocular depth estimation techniques (DA3~\cite{lin2025depth}, VGGT~\cite{wang2025vggt}), as well as state-of-the-art stereo models like FoundationStereo~\cite{wen2025foundationstereo} and S2M2~\cite{min2025s2m2}, under identical conditions.

To evaluate monocular depth capabilities, we first convert the predicted disparity into depth. We then perform an alignment by calculating the scale and shift via a least-squares optimization, using the ground-truth depth as a reference. The resulting scaled depth is used to compute standard monocular depth metrics: Absolute Relative Error ($Abs_{rel}$), and the threshold accuracy $\delta_1$.

We evaluated our model on the ETH3D~\cite{schoeps2017cvpr}, SimpleProc-S~\cite{ma2026fullyproceduralsyntheticdata}, and XYZ-IBD~\cite{huang2025xyzibdhighprecisionbinpickingdataset}. As reported in Tables \ref{tab:monocular_results} and \ref{tab:stereo_results}, our proposed method achieves highly competitive performance in monocular depth estimation task while maintaining SoTA results on stereo depth estimation. Notably, our approach narrows the performance gap between dedicated monocular estimators and stereo-based methods.

\begin{figure}[H]
  \centering
  \includegraphics[width=\linewidth]{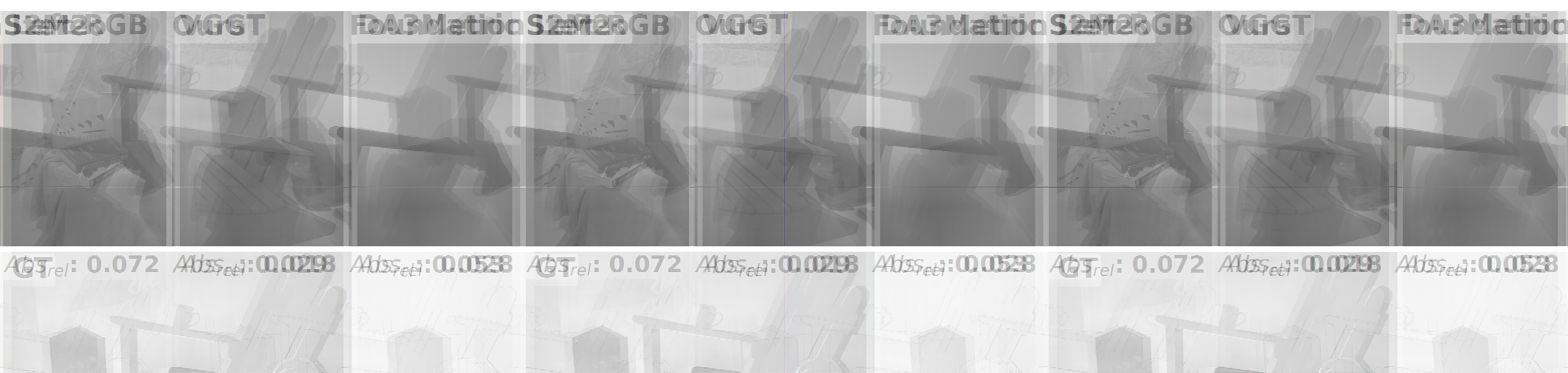}
  \caption{Our model (Col. 6) demonstrates superior visual quality for monocular depth when compared to SoTA stereo methods S2M2~\cite{min2025s2m2} (Col. 4) and FS~\cite{wen2025foundationstereo} (col 5) while being competitive with SoTA monodepth models DA3~\cite{lin2025depth} (Col. 3) and VGGT~\cite{wang2025vggt} (Col. 2).}
  \label{fig:monodepth}
\end{figure}

Our experiments reveal a trade-off in existing architectures: models trained primarily for monocular depth often underperform on stereo tasks, while leading stereo models typically struggle when restricted to monocular inputs (see Table \ref{tab:stereo_results}). In contrast, our model preserves SoTA stereo precision while remaining highly competitive in monocular scenarios ranking among the top 2 models for $Abs_{rel}$ and $\delta_1$ across multiple datasets. As illustrated in Figure \ref{fig:monodepth}, the model produces physically plausible disparity maps even under total right-camera occlusion, suggesting it has developed a robust internal representation of geometric priors.
\begin{figure}[h]
  \centering
  
  \centering
  \begin{minipage}{0.48\textwidth}
    \centering
    
    \captionof{table}{Our monocular depth estimation rivals DA3~\cite{lin2025depth} and VGGT~\cite{wang2025vggt} while outperforming sister stereo method S2M2~\cite{min2025s2m2}. FS~\cite{wen2025foundationstereo} uses a frozen DA2~\cite{kang2024depthanythingv2} trained for mono-depth.
    }
    \resizebox{\textwidth}{!}{%
\begin{tabular}{l cccc cc}
\toprule
\multirow{2}{*}{Model} & \multicolumn{2}{c}{ETH3D~\cite{eth3d_stereo}} & \multicolumn{2}{c}{SimpleProc-S~\cite{ma2026fullyproceduralsyntheticdata}} & \multicolumn{2}{c}{XYZ-IBD~\cite{huang2025xyzibdhighprecisionbinpickingdataset}} \\ 
\cmidrule(lr){2-3} \cmidrule(lr){4-5} \cmidrule(lr){6-7}
 & Abs\_rel $\downarrow$ & $\delta_1$ $\uparrow$ & Abs\_rel $\downarrow$ & $\delta_1$ $\uparrow$ & Abs\_rel $\downarrow$ & $\delta_1$ $\uparrow$ \\ 
\midrule
DA3Large~\cite{lin2025depth} & 0.108 & 0.896 & 0.708 & 0.614 & 3.413 & 0.137 \\ 
DA3Metric~\cite{lin2025depth} & \textit{0.068}\textsuperscript{2} & \textit{0.946}\textsuperscript{2} & 0.699 & 0.641 & 3.383 & 0.127 \\ 
VGGT~\cite{wang2025vggt} & \textbf{0.065}\textsuperscript{1} & \textbf{0.956}\textsuperscript{1} & 0.694 & \textbf{0.685}\textsuperscript{1} & 3.624 & 0.140 \\ 
\midrule
S2M2~\cite{min2025s2m2} & 0.220 & 0.666 & 0.824 & 0.569 & 3.389 & \textit{0.144}\textsuperscript{2} \\ 
FS~\cite{wen2025foundationstereo} (DA2~\cite{kang2024depthanythingv2}) & 0.160 & 0.790 & \textit{0.586}\textsuperscript{2} & \textit{0.643}\textsuperscript{2} & \textbf{2.700}\textsuperscript{1} & 0.140 \\ 
\midrule
\textbf{Ours} & 0.164 & 0.773 & \textbf{0.470}\textsuperscript{1} & 0.628 & \textit{2.788}\textsuperscript{2} & \textbf{0.172}\textsuperscript{1} \\ 
\bottomrule
\end{tabular}
    }
    \label{tab:monocular_results}
    
    
  \end{minipage}
  \hfill
  \begin{minipage}{0.48\textwidth}
    \captionof{table}{When evaluating stereo depth estimation instead of disparity estimation, our model still achieves state-of-the-art results across most metrics and benchmarks.}
    \resizebox{\textwidth}{!}{%
\begin{tabular}{l cccc cc}
\toprule
\multirow{2}{*}{Model} & \multicolumn{2}{c}{ETH3D~\cite{eth3d_stereo}} & \multicolumn{2}{c}{SimpleProc-S~\cite{ma2026fullyproceduralsyntheticdata}} & \multicolumn{2}{c}{XYZ-IBD~\cite{huang2025xyzibdhighprecisionbinpickingdataset}} \\ 
\cmidrule(lr){2-3} \cmidrule(lr){4-5} \cmidrule(lr){6-7}
 & Abs\_rel $\downarrow$ & $\delta_1$ $\uparrow$ & Abs\_rel $\downarrow$ & $\delta_1$ $\uparrow$ & Abs\_rel $\downarrow$ & $\delta_1$ $\uparrow$ \\ 
\midrule
DA3Large~\cite{lin2025depth} & 2269.640 & 0.000 & 3357.060 & 0.000 & 991.903 & 0.000 \\ 
DA3Metric~\cite{lin2025depth} & 0.238 & 0.814 & 4.640 & 0.175 & 1.601 & 0.035 \\ 
VGGT~\cite{wang2025vggt} & 0.299 & 0.967 & 79.935 & 0.001 & 0.862 & 0.000 \\ 
\midrule
S2M2~\cite{min2025s2m2} & 0.035 & \textit{0.995}\textsuperscript{2} & \textbf{0.026}\textsuperscript{1} & \textit{0.987}\textsuperscript{2} & 0.203 & \textit{0.813}\textsuperscript{2} \\ 
FS~\cite{wen2025foundationstereo} & \textit{0.028}\textsuperscript{2} & \textit{0.995}\textsuperscript{2} & 0.059 & 0.977 & \textit{0.132}\textsuperscript{2} & 0.830 \\ 
\midrule
\textbf{Ours} & \textbf{0.021}\textsuperscript{1} & \textbf{0.998}\textsuperscript{1} & \textit{0.030}\textsuperscript{2} & \textbf{0.989}\textsuperscript{1} & \textbf{0.128}\textsuperscript{1} & \textbf{0.882}\textsuperscript{1} \\ 
\bottomrule
\end{tabular}
    } 
    \label{tab:stereo_results}
  \end{minipage}

\end{figure}

\subsection{Comparison with 3D Foundation Models}
Architecturally, our approach shares similarities with recent 3D foundation models, including Depth Anything 3 \cite{lin2025depth} and VGGT \cite{wang2025vggt}. However, the primary distinction lies in our output representation. To provide a fair comparison across these different paradigms, In Table~\ref{tab:stereo_results} we evaluate all models on depth-based metrics by converting our predicted disparity maps into absolute depth using camera parameters. The results highlight our model's competitive performance in both zero-shot generalization and high-frequency detail preservation compared to these broad-purpose foundation models.

\subsection{Positional Embedding Spatial Resolution Analysis}
\label{subsec:upsampling_pe}
We evaluate the impact of different positional embedding spatial resolutions on disparity estimation. For this ablation, we trained a model exclusively on the FSD dataset~\cite{wen2025foundationstereo} using two spatial resolutions: $37\times37$ and $148\times148$. Performance was measured on a random 1\% validation split from FSD.

Quantitatively, the high-capacity PE grid yields robust performance gains. When inferring at a $966 \times 546$ scale, the $148 \times 148$ grid reduced the strict bad@0.5 error by an absolute 7.21 points compared to the $37 \times 37$ baseline. Similar error reductions were observed at the $280 \times 518$ scale (bad@0.5 reduced by 3.83 points), confirming that high-resolution spatial priors are robust and universally beneficial across varying inference dimensions.

\begin{figure}[H]
    \centering
    \setlength{\tabcolsep}{3pt} 
    \begin{tabular}{cccc}
        & Cosine Similarity & PCA Projection & First Channel \\[1ex] 
        
        \rotatebox[origin=c]{90}{\textbf{$37 \times 37$}} &
        \raisebox{-0.5\height}{\includegraphics[width=0.31\textwidth]{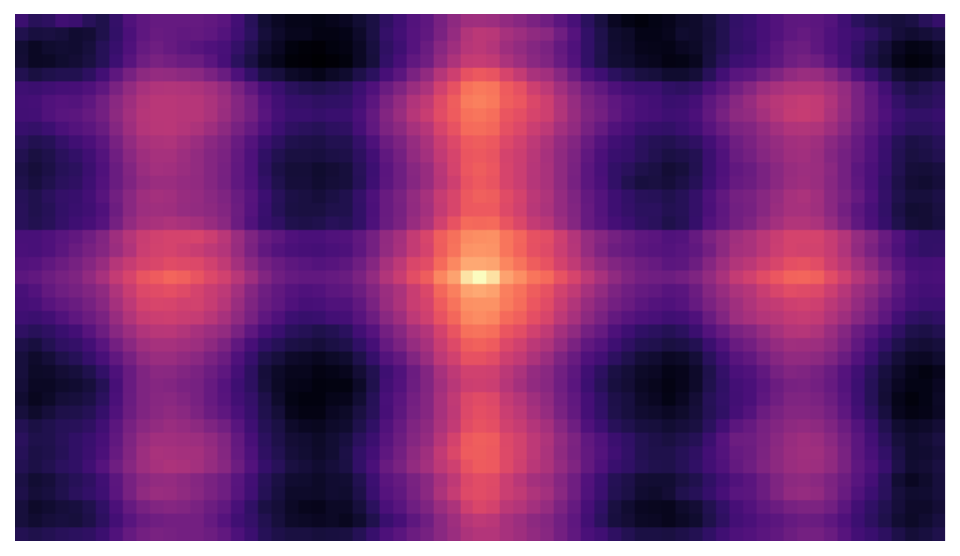}} &
        \raisebox{-0.5\height}{\includegraphics[width=0.31\textwidth]{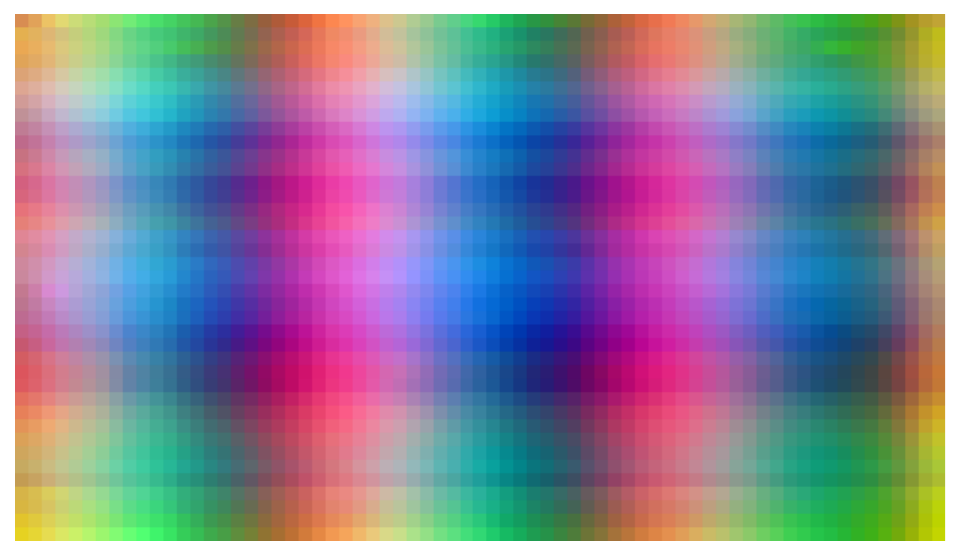}} &
        \raisebox{-0.5\height}{\includegraphics[width=0.31\textwidth]{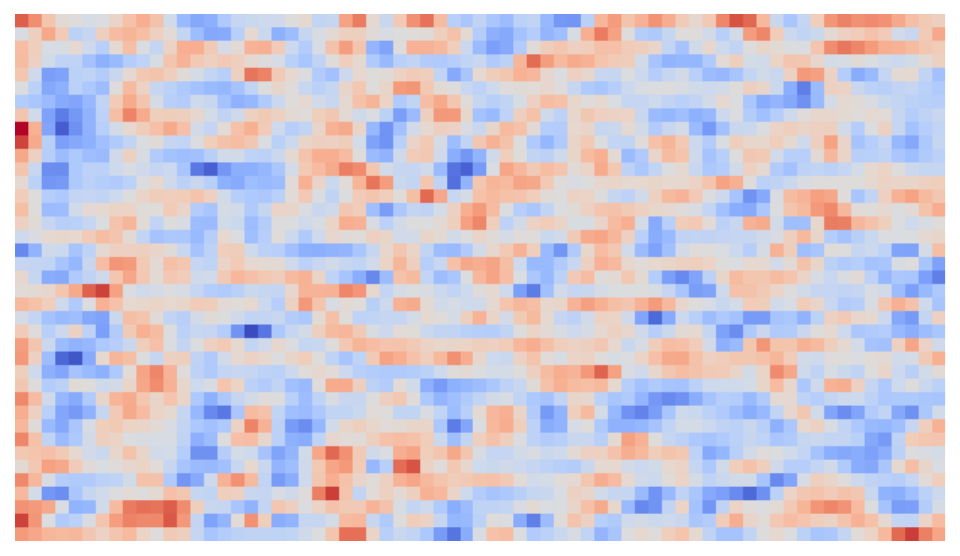}} \\[0.5cm] 
        
        \rotatebox[origin=c]{90}{\textbf{$148 \times 148$}} &
        \raisebox{-0.5\height}{\includegraphics[width=0.31\textwidth]{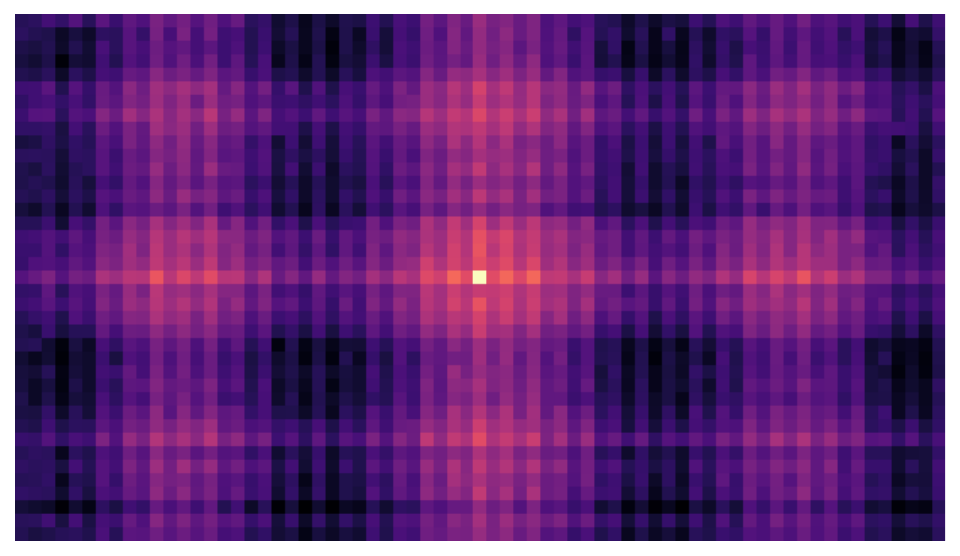}} &
        \raisebox{-0.5\height}{\includegraphics[width=0.31\textwidth]{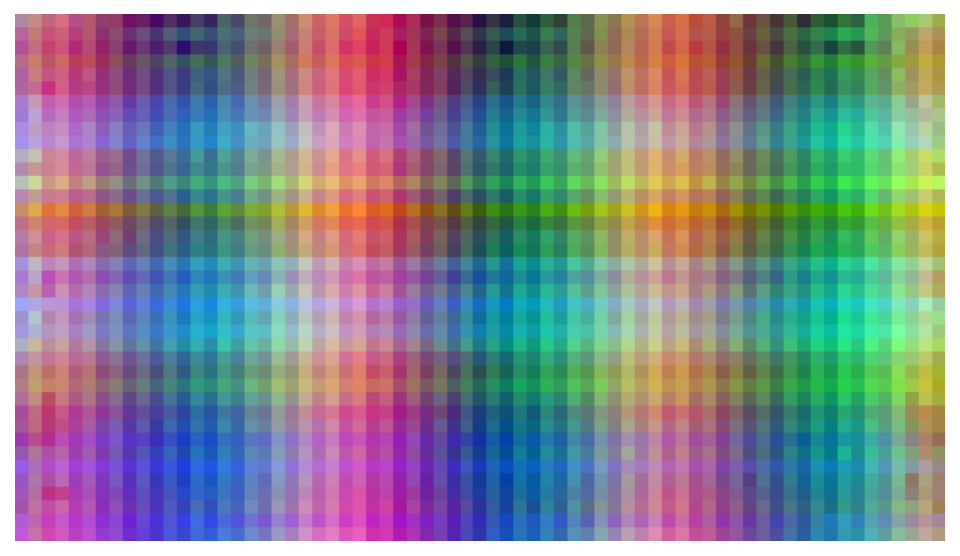}} &
        \raisebox{-0.5\height}{\includegraphics[width=0.31\textwidth]{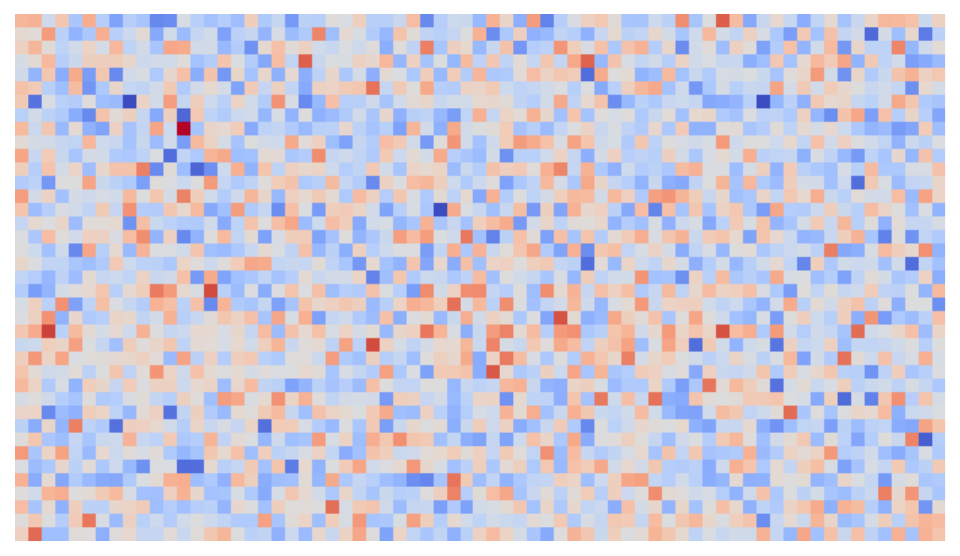}} \\
    \end{tabular}
    \caption{As the model trains, the learned $148 \times 148$ positional embeddings (Bottom) become significantly sharper spatial features than $37 \times 37$ (Top), which in turn allows significantly better disparity estimates. We visualize the cosine similarity with the center pixel (Left), the PCA projection (Middle), and the weight of the first channel (Right).}
    \label{fig:pe}
\end{figure}

Qualitatively (Figure~\ref{fig:pe}), the $37 \times 37$ embeddings successfully capture low-frequency spatial priors, but are insufficient to accurately estimate half-pixel disparities on a 966 pixel grid. The  $148 \times 148$ positional embeddings allow the network to learn refined, high-frequency spatial variations. This density provides the "spatial real estate" necessary to encode localized coordinate systems and resolve highly textured regions with enhanced precision.

\section{Limitations}
\label{sec:limitations}

While our inductive-bias-free architecture demonstrates strong performance, it currently exhibits a few limitations. First, the model is strictly trained to predict positive disparities; consequently, it cannot accurately resolve negative disparities, which can occasionally occur in real-world deployments. Second, because our final model weights are fine-tuned for specific, fixed spatial resolutions, input images must be resized or padded to match these exact dimensions prior to inference. This requirement restricts the model's flexibility compared to architectures capable of dynamically processing arbitrary image sizes and aspect ratios on the fly. Finally, similar to DA3~\cite{lin2025depth} and VGGT~\cite{wang2025vggt}, our reliance on a standard ViT backbone introduces a severe quadratic memory bottleneck during training at extreme resolutions. Because our architecture must be fine-tuned at a specific spatial dimension to natively process it during inference, we cannot currently evaluate ultra-high-resolution datasets such as Middlebury~\cite{middlebury_stereo} ($2872 \times 1984$) and Booster~\cite{zamaramirez2024booster} ($3840 \times 2160$) at their native scales. While highly specialized, fully convolutional or iterative stereo networks can dynamically process these extreme dimensions during inference without retraining, our current setup requires downsampling them to our maximum trained resolution ($1932 \times 1330$). Scaling our native training pipeline to accommodate these extreme dimensions remains an important area for future work. However, because our model relies entirely on general-purpose attention rather than custom-engineered geometry modules, it is perfectly positioned to benefit from the rapidly expanding ecosystem of Vision Transformer optimization techniques, such as token merging, efficient attention mechanisms, and quantization. 

\section{Conclusion}
In this work, we challenged the long-standing paradigm that explicit architectural inductive biases—such as 3D convolutions, cost volumes, and iterative refinement—are strictly necessary for state-of-the-art stereo reconstruction. By scaling a simple, end-to-end Vision Transformer on a massive synthetic dataset, we demonstrated that pure data-driven learning can match and surpass explicitly engineered geometry. Our model achieves highly competitive accuracy across standard benchmarks and naturally learns to fuse inter-image matching with intra-image monocular priors without requiring articulated sub-modules. We hope this work serves as a stepping stone, encouraging the community to move beyond highly specialized stereo pipelines and triggering further research into unified, general-purpose architectures for 3D reconstruction.

\bibliographystyle{unsrtnat}
\bibliography{main.bib}

@inproceedings{wen2025foundationstereo,
  title={Foundationstereo: Zero-shot stereo matching},
  author={Wen, Bowen and Trepte, Matthew and Aribido, Joseph and Kautz, Jan and Gallo, Orazio and Birchfield, Stan},
  booktitle={Proceedings of the IEEE/CVF conference on computer vision and pattern recognition},
  pages={5249--5260},
  year={2025}
}

@inproceedings{ranftl2021dpt,
  title={Vision Transformers for Dense Prediction},
  author={Ranftl, Ren{\'e} and Bochkovskiy, Alexey and Koltun, Vladlen},
  booktitle={Proceedings of the IEEE/CVF International Conference on Computer Vision (ICCV)},
  pages={12171--12182},
  year={2021}
}

@article{hirschmuller2008stereo,
  title={Stereo processing by semi-global matching and mutual information},
  author={Hirschmuller, Heiko},
  journal={IEEE Transactions on pattern analysis and machine intelligence},
  volume={30},
  number={2},
  pages={328--341},
  year={2008},
  publisher={IEEE}
}

@inproceedings{kendall2017end,
  title={End-to-end learning of geometry and context for deep stereo regression},
  author={Kendall, Alex and Martirosyan, Hayk and Dasgupta, Saumitro and Henry, Peter and Kennedy, Ryan and Bachrach, Abraham and Bry, Adam},
  booktitle={Proceedings of the IEEE international conference on computer vision},
  pages={66--75},
  year={2017}
}

@inproceedings{duggal2019deeppruner,
  title={DeepPruner: Learning efficient stereo matching via differentiable patchmatch},
  author={Duggal, Shivam and Wang, Shenlong and Ma, Wei-Chiu and Hu, Rui and Urtasun, Raquel},
  booktitle={Proceedings of the IEEE/CVF international conference on computer vision},
  pages={4384--4393},
  year={2019}
}

@inproceedings{chang2018pyramid,
  title={Pyramid stereo matching network},
  author={Chang, Jia-Ren and Chen, Yong-Sheng},
  booktitle={Proceedings of the IEEE conference on computer vision and pattern recognition},
  pages={5410--5418},
  year={2018}
}

@inproceedings{xu2020aanet,
  title={Aanet: Adaptive aggregation network for efficient stereo matching},
  author={Xu, Haofei and Zhang, Juyong},
  booktitle={Proceedings of the IEEE/CVF conference on computer vision and pattern recognition},
  pages={1959--1968},
  year={2020}
}

@inproceedings{shen2021cfnet,
  title={CFNet: Cascade and fused cost volume for robust stereo matching},
  author={Shen, Zhelun and Dai, Yuchao and Rao, Zhibo},
  booktitle={Proceedings of the IEEE/CVF Conference on Computer Vision and Pattern Recognition},
  pages={13906--13915},
  year={2021}
}

@inproceedings{shen2022pcw,
  title={PCW-Net: Pyramid combination and warping cost volume for stereo matching},
  author={Shen, Zhelun and Dai, Yuchao and Song, Xibin and Rao, Zhibo and Zhou, Dingfu and Zhang, Liangjun},
  booktitle={European Conference on Computer Vision},
  pages={280--297},
  year={2022},
  organization={Springer}
}

@inproceedings{lipson2021raft,
  title={Raft-stereo: Multilevel recurrent field transforms for stereo matching},
  author={Lipson, Lahav and Teed, Zachary and Deng, Jia},
  booktitle={International Conference on 3D Vision (3DV)},
  pages={218--227},
  year={2021},
  organization={IEEE}
}

@inproceedings{li2022practical,
  title={Practical stereo matching via cascaded recurrent network with adaptive correlation},
  author={Li, Jiankun and Wang, Peisen and Xiong, Pengfei and Cai, Tao and Yan, Ziwei and Yang, Lei and Liu, Jiangyu and Fan, Haoqiang and Liu, Shuaicheng},
  booktitle={Proceedings of the IEEE/CVF Conference on Computer Vision and Pattern Recognition},
  pages={16263--16272},
  year={2022}
}

@inproceedings{xu2023iterative,
  title={Iterative geometry encoding volume for stereo matching},
  author={Xu, Gangwei and Wang, Xianqi and Ding, Xiaohuan and Yang, Xin},
  booktitle={Proceedings of the IEEE/CVF Conference on Computer Vision and Pattern Recognition},
  pages={21919--21928},
  year={2023}
}

@article{xu2024igev++,
  title={IGEV++: Iterative multi-range geometry encoding volumes for stereo matching},
  author={Xu, Gangwei and Wang, Xianqi and Zhang, Zhaoxing and Cheng, Junda and Liao, Chunyuan and Yang, Xin},
  journal={arXiv preprint arXiv:2409.00638},
  year={2024}
}

@inproceedings{wang2024selective,
  title={Selective-stereo: Adaptive frequency information selection for stereo matching},
  author={Wang, Xianqi and Xu, Gangwei and Jia, Hao and Yang, Xin},
  booktitle={Proceedings of the IEEE/CVF Conference on Computer Vision and Pattern Recognition},
  pages={19701--19710},
  year={2024}
}

@inproceedings{chen2024mocha,
  title={Mocha-stereo: Motif channel attention network for stereo matching},
  author={Chen, Ziyang and Long, Wei and Yao, He and Zhang, Yongjun and Wang, Bingshu and Qin, Yongbin and Wu, Jia},
  booktitle={Proceedings of the IEEE/CVF Conference on Computer Vision and Pattern Recognition},
  pages={27768--27777},
  year={2024}
}

@article{tosi2024neural,
  title={Neural disparity refinement},
  author={Tosi, Fabio and Aleotti, Filippo and Ramirez, Pierluigi Zama and Poggi, Matteo and Salti, Samuele and Mattoccia, Stefano and Di Stefano, Luigi},
  journal={IEEE Transactions on Pattern Analysis and Machine Intelligence},
  year={2024},
  publisher={IEEE}
}

@inproceedings{bangunharcana2021correlate,
  title={Correlate-and-excite: Real-time stereo matching via guided cost volume excitation},
  author={Bangunharcana, Antyanta and Cho, Jae Won and Lee, Seokju and Kweon, In So and Kim, Kyung-Soo and Kim, Soohyun},
  booktitle={IEEE/RSJ International Conference on Intelligent Robots and Systems (IROS)},
  pages={3542--3548},
  year={2021},
  organization={IEEE}
}

@article{xu2023accurate,
  title={Accurate and efficient stereo matching via attention concatenation volume},
  author={Xu, Gangwei and Wang, Yun and Cheng, Junda and Tang, Jinhui and Yang, Xin},
  journal={IEEE Transactions on Pattern Analysis and Machine Intelligence},
  volume={46},
  number={4},
  pages={2461--2474},
  year={2023},
  publisher={IEEE}
}

@inproceedings{xu2021bilateral,
  title={Bilateral grid learning for stereo matching networks},
  author={Xu, Bin and Xu, Yuhua and Yang, Xiaoli and Jia, Wei and Guo, Yulan},
  booktitle={Proceedings of the IEEE/CVF conference on computer vision and pattern recognition},
  pages={12497--12506},
  year={2021}
}

@article{xu2023unifying,
  title={Unifying flow, stereo and depth estimation},
  author={Xu, Haofei and Zhang, Jing and Cai, Jianfei and Rezatofighi, Hamid and Yu, Fisher and Tao, Dacheng and Geiger, Andreas},
  journal={IEEE Transactions on Pattern Analysis and Machine Intelligence},
  year={2023},
  publisher={IEEE}
}

@inproceedings{bartolomei2024stereo,
  title={Stereo anywhere: Robust zero-shot deep stereo matching even where either stereo or mono fail},
  author={Bartolomei, Luca and Tosi, Fabio and Poggi, Matteo and Mattoccia, Stefano},
  booktitle={Proceedings of the IEEE/CVF Conference on Computer Vision and Pattern Recognition},
  pages={1013--1027},
  year={2025}
}

@inproceedings{jiang2025defom,
  title={Defom-stereo: Depth foundation model based stereo matching},
  author={Jiang, Hualie and Lou, Zhiqiang and Ding, Laiyan and Xu, Rui and Tan, Minglang and Jiang, Wenjie and Huang, Rui},
  booktitle={Proceedings of the IEEE/CVF Conference on Computer Vision and Pattern Recognition},
  pages={21857--21867},
  year={2025}
}

@inproceedings{zhou2025all,
  title={All-in-one: Transferring vision foundation models into stereo matching},
  author={Zhou, Jingyi and Zhang, Haoyu and Yuan, Jiakang and Ye, Peng and Chen, Tao and Jiang, Hao and Chen, Meiya and Zhang, Yangyang},
  booktitle={Proceedings of the AAAI Conference on Artificial Intelligence},
  pages={10797--10805},
  year={2025}
}

@inproceedings{wang2024dust3r,
  title={Dust3r: Geometric 3d vision made easy},
  author={Wang, Shuzhe and Leroy, Vincent and Cabon, Yohann and Chidlovskii, Boris and Revaud, Jerome},
  booktitle={Proceedings of the IEEE/CVF Conference on Computer Vision and Pattern Recognition},
  pages={20697--20709},
  year={2024}
}

@inproceedings{wang2025vggt,
  title={Vggt: Visual geometry grounded transformer},
  author={Wang, Jianyuan and Chen, Minghao and Karaev, Nikita and Vedaldi, Andrea and Rupprecht, Christian and Novotny, David},
  booktitle={Proceedings of the IEEE/CVF Conference on Computer Vision and Pattern Recognition},
  pages={5294--5306},
  year={2025}
}

@inproceedings{yang2025fast3r,
  title={Fast3r: Towards 3d reconstruction of 1000+ images in one forward pass},
  author={Yang, Jianing and Sax, Alexander and Liang, Kevin J and Henaff, Mikael and Tang, Hao and Cao, Ang and Chai, Joyce and Meier, Franziska and Feiszli, Matt},
  booktitle={Proceedings of the IEEE/CVF Conference on Computer Vision and Pattern Recognition},
  pages={21924--21935},
  year={2025}
}

@article{kang2024depthanythingv2,
  title={Depth Anything V2},
  author={Yang, Lihe and Kang, Bingyi and Huang, Zilong and Zhao, Zhen and Xu, Xiaogang and Feng, Jiashi and Zhao, Hengshuang},
  journal={arXiv preprint arXiv:2406.09414},
  year={2024}
}

@article{hu2024metric3dv2,
  title={Metric3Dv2: A Versatile Monocular Geometric Foundation Model for Zero-shot Metric Depth and Surface Normal Estimation},
  author={Hu, Mu and Yin, Wei and Zhang, Chi and Cai, Zhipeng and Long, Xiaoxiao and Wang, Kaixuan and Chen, Hao and Yu, Gang and Shen, Chunhua and Shen, Shaojie},
  journal={IEEE Transactions on Pattern Analysis and Machine Intelligence},
  year={2024},
  publisher={IEEE}
}

@inproceedings{bochkovskiy2024depthpro,
  title={Depth Pro: Sharp Monocular Metric Depth in Less Than a Second},
  author={Bochkovskii, Aleksei and Delaunoy, Ama{\"e}l and Germain, Hugo and Santos, Marcel and Zhou, Yichao and Richter, Stephan R. and Koltun, Vladlen},
  booktitle={International Conference on Learning Representations (ICLR)},
  year={2025}
}

@inproceedings{wang2024moge,
  title={Moge: Unlocking accurate monocular geometry estimation for open-domain images with optimal training supervision},
  author={Wang, Ruicheng and Xu, Sicheng and Dai, Cassie and Xiang, Jianfeng and Deng, Yu and Tong, Xin and Yang, Jiaolong},
  booktitle={Proceedings of the Computer Vision and Pattern Recognition Conference},
  pages={5261--5271},
  year={2025}
}

@inproceedings{wang2025moge2,
 author = {Wang, Ruicheng and Xu, Sicheng and Dong, Yue and Deng, Yu and Xiang, Jianfeng and Lv, Zelong and Sun, Guangzhong and Tong, Xin and Yang, Jiaolong},
 booktitle = {Advances in Neural Information Processing Systems},
 doi = {10.52202/085713-1207},
 editor = {D. Belgrave and C. Zhang and H. Lin and R. Pascanu and P. Koniusz and M. Ghassemi and N. Chen},
 pages = {35928--35959},
 publisher = {Curran Associates, Inc.},
 title = {MoGe-2: Accurate Monocular Geometry with Metric Scale and Sharp Details},
 url = {https://proceedings.neurips.cc/paper_files/paper/2025/file/336572db3e99930814d6b328d4220cb6-Paper-Conference.pdf},
 volume = {38, Main Conference},
 year = {2025}
}

@article{wen2025fastfoundationstereo,
  title={Fast-FoundationStereo: Real-Time Zero-Shot Stereo Matching},
  author={Wen, Bowen and Dewan, Shaurya and Birchfield, Stan},
  journal={arXiv preprint arXiv:2512.11130},
  year={2025}
}

@article{dosovitskiy2020image,
  title={An image is worth 16x16 words: Transformers for image recognition at scale},
  author={Dosovitskiy, Alexey and Beyer, Lucas and Kolesnikov, Alexander and Weissenborn, Dirk and Zhai, Xiaohua and Unterthiner, Thomas and Dehghani, Mostafa and Minderer, Matthias and Heigold, Georg and Gelly, Sylvain and others},
  journal={arXiv preprint arXiv:2010.11929},
  year={2020}
}

@article{simeoni2025dinov3,
  title={Dinov3},
  author={Sim{\'e}oni, Oriane and Vo, Huy V and Seitzer, Maximilian and Baldassarre, Federico and Oquab, Maxime and Jose, Cijo and Khalidov, Vasil and Szafraniec, Marc and Yi, Seungeun and Ramamonjisoa, Micha{\"e}l and others},
  journal={arXiv preprint arXiv:2508.10104},
  year={2025}
}

@article{lin2025depth,
  title={Depth anything 3: Recovering the visual space from any views},
  author={Lin, Haotong and Chen, Sili and Liew, Junhao and Chen, Donny Y and Li, Zhenyu and Shi, Guang and Feng, Jiashi and Kang, Bingyi},
  journal={arXiv preprint arXiv:2511.10647},
  year={2025}
}

@inproceedings{li2021revisiting,
  title={Revisiting stereo depth estimation from a sequence-to-sequence perspective with transformers},
  author={Li, Zhaoshuo and Liu, Xingtong and Drenkow, Nathan and Ding, Andy and Creighton, Francis X and Taylor, Russell H and Unberath, Mathias},
  booktitle={Proceedings of the IEEE/CVF international conference on computer vision},
  pages={6197--6206},
  year={2021}
}

@inproceedings{weinzaepfel2023croco,
  title={Croco v2: Improved cross-view completion pre-training for stereo matching and optical flow},
  author={Weinzaepfel, Philippe and Lucas, Thomas and Leroy, Vincent and Cabon, Yohann and Arora, Vaibhav and Br{\'e}gier, Romain and Csurka, Gabriela and Antsfeld, Leonid and Chidlovskii, Boris and Revaud, J{\'e}r{\^o}me},
  booktitle={Proceedings of the IEEE/CVF International Conference on Computer Vision},
  pages={17969--17980},
  year={2023}
}

@article{kim2025unitt,
  title={Unitt-stereo: Unified training of transformer for enhanced stereo matching},
  author={Kim, Soomin and Choi, Hyesong and Ahn, Jihye and Min, Dongbo},
  journal={IEEE Access},
  volume={13},
  pages={204695--204707},
  year={2025},
  publisher={IEEE}
}

@inproceedings{min2025s2m2,
  title={S2M2: Scalable Stereo Matching Model for Reliable Depth Estimation},
  author={Min, Junhong and Jeon, Youngpil and Kim, Jimin and Choi, Minyong},
  booktitle={Proceedings of the IEEE/CVF International Conference on Computer Vision},
  pages={26729--26739},
  year={2025}
}

@inproceedings{guo2022context,
  title={Context-enhanced stereo transformer},
  author={Guo, Weiyu and Li, Zhaoshuo and Yang, Yongkui and Wang, Zheng and Taylor, Russell H and Unberath, Mathias and Yuille, Alan and Li, Yingwei},
  booktitle={European Conference on Computer Vision},
  pages={263--279},
  year={2022},
  organization={Springer}
}

@article{jia2024transformer,
  title={A transformer-based architecture for high-resolution stereo matching},
  author={Jia, Di and Cai, Peng and Wang, Qian and Yang, Ninghua},
  journal={IEEE Transactions on Computational Imaging},
  volume={10},
  pages={83--92},
  year={2024},
  publisher={IEEE}
}

@inproceedings{bengana2022seeking,
  title={Seeking attention: Using full context transformers for better disparity estimation},
  author={Bengana, Nadir and Mustaniemi, Janne and Heikkil{\"a}, Janne},
  booktitle={International Conference on Pattern Recognition and Artificial Intelligence},
  pages={398--409},
  year={2022},
  organization={Springer}
}

@misc{ma2026fullyproceduralsyntheticdata,
      title={Fully Procedural Synthetic Data from Simple Rules for Multi-View Stereo}, 
      author={Zeyu Ma and Alexander Raistrick and Jia Deng},
      year={2026},
      eprint={2604.04925},
      archivePrefix={arXiv},
      primaryClass={cs.CV},
      url={https://arxiv.org/abs/2604.04925}, 
}

@article{ranftl2020towards,
  title={Towards robust monocular depth estimation: Mixing datasets for zero-shot cross-dataset transfer},
  author={Ranftl, Ren{\'e} and Lasinger, Katrin and Hafner, David and Schindler, Konrad and Koltun, Vladlen},
  journal={IEEE transactions on pattern analysis and machine intelligence},
  volume={44},
  number={3},
  pages={1623--1637},
  year={2020},
  publisher={IEEE}
}

@inproceedings{li2018megadepth,
  title={Megadepth: Learning single-view depth prediction from internet photos},
  author={Li, Zhengqi and Snavely, Noah},
  booktitle={Proceedings of the IEEE conference on computer vision and pattern recognition},
  pages={2041--2050},
  year={2018}
}

@article{oquab2024dinov,
title={{DINO}v2: Learning Robust Visual Features without Supervision},
author={Maxime Oquab and Timoth{\'e}e Darcet and Th{\'e}o Moutakanni and Huy V. Vo and Marc Szafraniec and Vasil Khalidov and Pierre Fernandez and Daniel HAZIZA and Francisco Massa and Alaaeldin El-Nouby and Mido Assran and Nicolas Ballas and Wojciech Galuba and Russell Howes and Po-Yao Huang and Shang-Wen Li and Ishan Misra and Michael Rabbat and Vasu Sharma and Gabriel Synnaeve and Hu Xu and Herve Jegou and Julien Mairal and Patrick Labatut and Armand Joulin and Piotr Bojanowski},
journal={Transactions on Machine Learning Research},
issn={2835-8856},
year={2024},
}

@article{wang2026waft,
  title={WAFT-Stereo: Warping-Alone Field Transforms for Stereo Matching},
  author={Wang, Yihan and Deng, Jia},
  journal={arXiv preprint arXiv:2603.24836},
  year={2026}
}

@article{min2025depthfocus,
  title={Depthfocus: Controllable depth estimation for see-through scenes},
  author={Min, Junhong and Kim, Jimin and Kim, Minwook and Min, Cheol-Hui and Jeon, Youngpil and Choi, Minyong},
  journal={arXiv preprint arXiv:2511.16993},
  year={2025}
}

@article{scharstein2002taxonomy,
  title={A taxonomy and evaluation of dense two-frame stereo correspondence algorithms},
  author={Scharstein, Daniel and Szeliski, Richard},
  journal={International journal of computer vision},
  volume={47},
  number={1},
  pages={7--42},
  year={2002},
  publisher={Springer}
}

@misc{middlebury_stereo,
  title = {Middlebury Stereo Benchmark Page},
  howpublished = {\url{http://vision.middlebury.edu/stereo/}},
}

@inproceedings{scharstein2014high,
  title={High-resolution stereo datasets with subpixel-accurate ground truth},
  author={Scharstein, Daniel and Hirschm{\"u}ller, Heiko and Kitajima, York and Krathwohl, Greg and Ne{\v{s}}i{\'c}, Nera and Wang, Xi and Westling, Porter},
  booktitle={German conference on pattern recognition},
  pages={31--42},
  year={2014},
  organization={Springer}
}

@inproceedings{schoeps2017cvpr,
  author = {Thomas Sch\"ops and Johannes L. Sch\"onberger and Silvano Galliani and Torsten Sattler and Konrad Schindler and Marc Pollefeys and Andreas Geiger},
  title = {A Multi-View Stereo Benchmark with High-Resolution Images and Multi-Camera Videos},
  booktitle = {Conference on Computer Vision and Pattern Recognition (CVPR)},
  year = {2017}
}

@misc{eth3d_stereo,
  title = {ETH3D Stereo Benchmark Page},
  howpublished = {\url{https://www.eth3d.net/low_res_two_view}},
}

@misc{huang2025xyzibdhighprecisionbinpickingdataset,
    title={XYZ-IBD: High-precision Bin-picking Dataset for Object 6D Pose Estimation Capturing Real-world Industrial Complexity}, 
    author={Junwen Huang and Jizhong Liang and Jiaqi Hu and Martin Sundermeyer and Peter KT Yu and Nassir Navab and Benjamin Busam},
    year={2025},
    eprint={2506.00599},
    archivePrefix={arXiv},
    primaryClass={cs.CV},
    url={https://arxiv.org/abs/2506.00599},
}

@misc{liu2025muonscalablellmtraining,
      title={Muon is Scalable for LLM Training}, 
      author={Jingyuan Liu and Jianlin Su and Xingcheng Yao and Zhejun Jiang and Guokun Lai and Yulun Du and Yidao Qin and Weixin Xu and Enzhe Lu and Junjie Yan and Yanru Chen and Huabin Zheng and Yibo Liu and Shaowei Liu and Bohong Yin and Weiran He and Han Zhu and Yuzhi Wang and Jianzhou Wang and Mengnan Dong and Zheng Zhang and Yongsheng Kang and Hao Zhang and Xinran Xu and Yutao Zhang and Yuxin Wu and Xinyu Zhou and Zhilin Yang},
      year={2025},
      eprint={2502.16982},
      archivePrefix={arXiv},
      primaryClass={cs.LG},
      url={https://arxiv.org/abs/2502.16982}, 
}

@misc{loshchilov2019decoupledweightdecayregularization,
      title={Decoupled Weight Decay Regularization}, 
      author={Ilya Loshchilov and Frank Hutter},
      year={2019},
      eprint={1711.05101},
      archivePrefix={arXiv},
      primaryClass={cs.LG},
      url={https://arxiv.org/abs/1711.05101}, 
}

@misc{tremblay2018fallingthingssyntheticdataset,
      title={Falling Things: A Synthetic Dataset for 3D Object Detection and Pose Estimation}, 
      author={Jonathan Tremblay and Thang To and Stan Birchfield},
      year={2018},
      eprint={1804.06534},
      archivePrefix={arXiv},
      primaryClass={cs.CV},
      url={https://arxiv.org/abs/1804.06534}, 
}

@article{bao2020instereo2k,
  title={Instereo2k: a large real dataset for stereo matching in indoor scenes},
  author={Bao, Wei and Wang, Wei and Xu, Yuhua and Guo, Yulan and Hong, Siyu and Zhang, Xiaohu},
  journal={Science China Information Sciences},
  volume={63},
  number={11},
  pages={212101},
  year={2020},
  publisher={Springer}
}

@article{jing2024match,
  title={Match Stereo Videos via Bidirectional Alignment},
  author={Junpeng Jing and Ye Mao and Anlan Qiu and Krystian Mikolajczyk},
  year={2024}
}

@inproceedings{mehl2023spring,
    title={Spring: A High-Resolution High-Detail Dataset and Benchmark for Scene Flow, Optical Flow and Stereo},
    author={Mehl, Lukas and Schmalfuss, Jenny and Jahedi, Azin and Nalivayko, Yaroslava and Bruhn, Andr{\'e}s},
    booktitle={Proc. IEEE/CVF Conference on Computer Vision and Pattern Recognition (CVPR)},
    pages={4981--4991},
    year={2023}
}

@inproceedings{tosi2021smd,
  title={Smd-nets: Stereo mixture density networks},
  author={Tosi, Fabio and Liao, Yiyi and Schmitt, Carolin and Geiger, Andreas},
  booktitle={Proceedings of the IEEE/CVF conference on computer vision and pattern recognition},
  pages={8942--8952},
  year={2021}
}

@InProceedings{MIFDB16,
  author    = "N. Mayer and E. Ilg and P. H{\"a}usser and P. Fischer and D. Cremers and A. Dosovitskiy and T. Brox",
  title     = "A Large Dataset to Train Convolutional Networks for Disparity, Optical Flow, and Scene Flow Estimation",
  booktitle = "IEEE International Conference on Computer Vision and Pattern Recognition (CVPR)",
  year      = "2016",
  note      = "arXiv:1512.02134",
  url       = "http://lmb.informatik.uni-freiburg.de/Publications/2016/MIFDB16"
}

@misc{cabon2020virtualkitti2,
      title={Virtual KITTI 2}, 
      author={Yohann Cabon and Naila Murray and Martin Humenberger},
      year={2020},
      eprint={2001.10773},
      archivePrefix={arXiv},
      primaryClass={cs.CV},
      url={https://arxiv.org/abs/2001.10773}, 
}

@inproceedings{10.1109/IROS45743.2020.9341801,
author = {Wang, Wenshan and Zhu, Delong and Wang, Xiangwei and Hu, Yaoyu and Qiu, Yuheng and Wang, Chen and Hu, Yafei and Kapoor, Ashish and Scherer, Sebastian},
title = {TartanAir: A Dataset to Push the Limits of Visual SLAM},
year = {2020},
publisher = {IEEE Press},
url = {https://doi.org/10.1109/IROS45743.2020.9341801},
doi = {10.1109/IROS45743.2020.9341801},
booktitle = {2020 IEEE/RSJ International Conference on Intelligent Robots and Systems (IROS)},
pages = {4909–4916},
numpages = {8},
location = {Las Vegas, NV, USA}
}

@article{patel2025tartanground,
      title={TartanGround: A Large-Scale Dataset for Ground Robot Perception and Navigation},
      author={Patel, Manthan and Yang, Fan and Qiu, Yuheng and Cadena, Cesar and Scherer, Sebastian and Hutter, Marco and Wang, Wenshan},
      journal={arXiv preprint arXiv:2505.10696},
      year={2025}}

@INPROCEEDINGS{9428423,
  author={Wang, Qiang and Zheng, Shizhen and Yan, Qingsong and Deng, Fei and Zhao, Kaiyong and Chu, Xiaowen},
  booktitle={2021 IEEE International Conference on Multimedia and Expo (ICME)}, 
  title={IRS: A Large Naturalistic Indoor Robotics Stereo Dataset to Train Deep Models for Disparity and Surface Normal Estimation}, 
  year={2021},
  volume={},
  number={},
  pages={1-6},
  doi={10.1109/ICME51207.2021.9428423}}

@misc{yan2025proceduraldatasetgenerationzeroshot,
      title={What Makes Good Synthetic Training Data for Zero-Shot Stereo Matching?}, 
      author={David Yan and Alexander Raistrick and Jia Deng},
      year={2025},
      eprint={2504.16930},
      archivePrefix={arXiv},
      primaryClass={cs.CV},
      url={https://arxiv.org/abs/2504.16930}, 
}

@article{dao2023flashattention,
  title={Flashattention-2: Faster attention with better parallelism and work partitioning},
  author={Dao, Tri},
  journal={arXiv preprint arXiv:2307.08691},
  year={2023}
}

@inproceedings{cheng2025monster,
  title={Monster: Marry monodepth to stereo unleashes power},
  author={Cheng, Junda and Liu, Longliang and Xu, Gangwei and Wang, Xianqi and Zhang, Zhaoxing and Deng, Yong and Zang, Jinliang and Chen, Yurui and Cai, Zhipeng and Yang, Xin},
  booktitle={Proceedings of the Computer Vision and Pattern Recognition Conference},
  pages={6273--6282},
  year={2025}
}

@article{zamaramirez2024booster,
    author={Ramirez, Pierluigi Zama and Costanzino, Alex and Tosi, Fabio and Poggi, Matteo and Salti, Samuele and Mattoccia, Stefano and Stefano, Luigi Di},
    journal={IEEE Transactions on Pattern Analysis and Machine Intelligence}, 
    title={Booster: A Benchmark for Depth From Images of Specular and Transparent Surfaces}, 
    year={2024},
    volume={46},
    number={1},
    pages={85-102},
    doi={10.1109/TPAMI.2023.3323858}
}


\newpage
\appendix


\section{Supplement materials}

\subsection{Full Evaluation Results}

In this section, we present comprehensive quantitative evaluations across four major stereo benchmarks: XYZ-IBD~\cite{huang2025xyzibdhighprecisionbinpickingdataset}, SimpleProc-S~\cite{ma2026fullyproceduralsyntheticdata}, SimpleProc-M~\cite{ma2026fullyproceduralsyntheticdata}, and ETH3D~\cite{eth3d_stereo}. Due to space constraints in the main text, the full tables are detailed here to provide a complete picture of our method's performance relative to recent state-of-the-art architectures.

Across all four benchmarks, our proposed method demonstrates state-of-the-art or highly competitive performance, exhibiting strong generalization capabilities across synthetic, real-world, indoor, and outdoor settings:

\begin{itemize}
    \item \textbf{XYZ-IBD} (Table~\ref{tab:bop_results_supp}): On this challenging dataset, our model achieves top-tier results. It secures the absolute best performance in End-Point Error (EPE), bad@2 (\textbf{36.64}) and bad@4 (\textbf{23.39}), while remaining highly competitive across all other error thresholds.
    \item \textbf{SimpleProc-S / SimpleProc-M} (Table~\ref{tab:simpleproc_results_supp}): Evaluating across different image resolutions demonstrates our method's multi-scale robustness. At the $966 \times 546$ resolution, our model captures the top spot in almost all the metrics, ranking second only on bad@0.5 metric behind FoundationStereo. At the higher $1932 \times 1092$ resolution, it maintains dominance, achieving the lowest EPE (\textbf{0.40}) and reducing the bad@4 error by nearly half results compared to the second best model.
    \item \textbf{ETH3D~\cite{eth3d_stereo}} (Table~\ref{tab:eth3d-results_supp}): Our framework establishes a new state-of-the-art across virtually all metrics on the ETH3D test set. Notably, for non-occluded pixels, it achieves an EPE of \textbf{0.09} and slashes the bad@0.5 metric down to \textbf{0.65} (nearly half the error rate of FoundationStereo~\cite{wen2025foundationstereo}). This exceptional sub-pixel accuracy is consistently maintained when evaluating all pixels as well.
\end{itemize}

\begin{table}[h]
\centering
\caption{Quantitative comparison on the XYZ-IBD dataset~\cite{huang2025xyzibdhighprecisionbinpickingdataset}. The second-best values are highlighted with a superscript 2.}
\label{tab:bop_results_supp}
\resizebox{0.6\textwidth}{!}{
\begin{tabular}{l cccccc}
\toprule
Model & EPE & RMS  & bad@0.5 & bad@1 & bad@2 & bad@4 \\ 
\midrule
CroCo~\cite{weinzaepfel2023croco}            & 34.98 & 47.77 & 91.59 & 84.94 & 76.69 & 67.95 \\ 
CREStereo~\cite{li2022practical}         & 15.92$^{2}$ & \textbf{34.36} & 77.65 & 62.28 & 46.96 & 36.04 \\ 
FoundationStereo~\cite{wen2025foundationstereo} & 15.75 & 35.87 & \textbf{72.63} & \textbf{54.69} & 38.43$^{2}$ & 28.35$^{2}$ \\ 
Selective-IGEV~\cite{wang2024selective}    & 17.48 & 35.44$^{2}$ & 76.98 & 60.92 & 45.53 & 34.98 \\ 
S2M2~\cite{min2025s2m2}              & 19.24 & 49.05 & 75.26$^{2}$ & 57.22 & 40.74 & 30.69 \\ 
\midrule
\textbf{Ours}    & \textbf{11.39} & 38.21 & 74.22 & 55.61$^{2}$ & \textbf{36.64} & \textbf{23.39} \\ 
\bottomrule
\end{tabular}%
}
\end{table}
\begin{table}[h]
\centering
\caption{Quantitative comparison on SimpleProc~\cite{ma2026fullyproceduralsyntheticdata} dataset at different resolutions on all-pixels. The second-best values are highlighted with a superscript 2.}
\label{tab:simpleproc_results_supp}
\resizebox{\textwidth}{!}{%
\begin{tabular}{l cccccc cccccc}
\toprule
\multirow{2}{*}{Model} & \multicolumn{6}{c}{SimpleProc-S} & \multicolumn{6}{c}{SimpleProc-M} \\ 
\cmidrule(lr){2-7} \cmidrule(lr){8-13}
 & EPE & RMS & bad@0.5 & bad@1.0 & bad@2.0 & bad@4.0 & EPE & RMS & bad@0.5 & bad@1.0 & bad@2.0 & bad@4.0 \\ 
\midrule
CREStereo~\cite{li2022practical}         & 0.33 & 1.30 & 8.32 & 4.72 & 2.58 & 1.43 & 0.64 & 2.70 & 12.87 & 8.03 & 4.89 & 2.75 \\ 
CroCo~\cite{weinzaepfel2023croco} & 0.45 & 1.43 & 14.69 & 6.94 & 3.50 & 1.82 & 0.91 & 3.11 & 17.14 & 9.58 & 5.83 & 3.75 \\ 
FoundationStereo~\cite{wen2025foundationstereo} & 0.48 & 2.50 & \textbf{7.23} & 4.25 & 2.79 & 1.95 & 0.61 & 2.82 & 9.49$^{2}$ & 6.27 & 4.47 & 3.22 \\ 
Selective-IGEV~\cite{wang2024selective}   & 0.52 & 2.15 & 11.73 & 6.13 & 3.46 & 2.07 & 0.75 & 3.36 & 13.49 & 8.16 & 5.13 & 3.19 \\ 
S2M2~\cite{min2025s2m2}                  & 0.29$^{2}$ & 1.00$^{2}$ & 7.97 & 4.16$^{2}$ & 2.20$^{2}$ & 1.12$^{2}$ & 0.44$^{2}$ & \textbf{1.73} & 9.69 & 6.04$^{2}$ & 3.83$^{2}$ & 2.31$^{2}$ \\ 
\midrule
\textbf{Ours} & \textbf{0.25} & \textbf{0.80} & 7.83$^{2}$ & \textbf{3.03} & \textbf{1.36} & \textbf{0.63} & \textbf{0.40} & 1.91$^{2}$ & \textbf{8.59} & \textbf{4.22} & \textbf{2.26} & \textbf{1.26} \\ 
\bottomrule
\end{tabular}%
}
\end{table}
\begin{table}[t]
\centering
\caption{Results on the test set of ETH3D~\cite{eth3d_stereo}. Our method achieves state-of-the-art performance across both non-occluded and all-pixel regions.}
\label{tab:eth3d-results_supp}
\resizebox{\textwidth}{!}{%
\begin{tabular}{l cccccc cccccc}
\toprule
\multirow{2}{*}{Model} & \multicolumn{6}{c}{Non-occluded pixels} & \multicolumn{6}{c}{All pixels} \\ 
\cmidrule(lr){2-7} \cmidrule(lr){8-13}
 & EPE & RMS & bad@0.5 & bad@1 & bad@2 & bad@4 & EPE & RMS & bad@0.5 & bad@1 & bad@2 & bad@4 \\ 
\midrule
RAFT-Stereo~\cite{lipson2021raft}     & 0.18 & 0.36 & 7.04 & 2.44 & 0.44 & 0.15 & 0.19 & 0.42 & 7.33 & 2.60 & 0.56 & 0.22 \\
CroCo~\cite{weinzaepfel2023croco}  & 0.14 & 0.30 & 3.27 & 0.99 & 0.39 & 0.13 & 0.15 & 0.35 & 3.51 & 1.14 & 0.50 & 0.18 \\
CREStereo~\cite{li2022practical}      & 0.13 & 0.28 & 3.58 & 0.98 & 0.22 & 0.10 & 0.14 & 0.31 & 3.75 & 1.09 & 0.29 & 0.12 \\
FoundationStereo~\cite{wen2025foundationstereo} & \textbf{0.09} & 0.20 & 1.26 & 0.26 & 0.08 & 0.05 & 0.13 & 0.61 & 1.56 & 0.48 & 0.26 & 0.21 \\
S2M2 XL~\cite{min2025s2m2}            & 0.10 & 0.20 & 0.93 & 0.22 & 0.06 & 0.03 & 0.10 & 0.22 & 1.03 & 0.26 & 0.08 & 0.04 \\
WAFT-Stereo~\cite{wang2026waft}       & 0.10 & 0.20 & 0.89 & 0.28 & 0.07 & 0.03 & 0.10 & 0.21 & 0.97 & 0.31 & 0.07 & 0.03 \\
DepthFocus~\cite{min2025depthfocus}   & 0.10 & 0.19 & 1.04 & 0.25 & 0.07 & \textbf{0.02} & 0.10 & 0.22 & 1.14 & 0.29 & 0.09 & 0.04 \\ 
\midrule
\textbf{Ours} & \textbf{0.09} & \textbf{0.17} & \textbf{0.65} & \textbf{0.14} & \textbf{0.04} & \textbf{0.02} & \textbf{0.09} & \textbf{0.18} & \textbf{0.71} & \textbf{0.16} & \textbf{0.04} & \textbf{0.02} \\ 
\bottomrule
\end{tabular}%
}
\end{table}

\subsection{Generating SimpleProc Evaluation Dataset}

\begin{figure}[htbp]
    \centering
    
    \renewcommand{\arraystretch}{1.5} 
    
    \setlength{\tabcolsep}{2pt}
    
    \begin{tabular}{@{} c c c c c @{}}
        
        \rotatebox[origin=c]{90}{\textbf{Left Image}} & 
        \includegraphics[width=0.22\linewidth, valign=m]{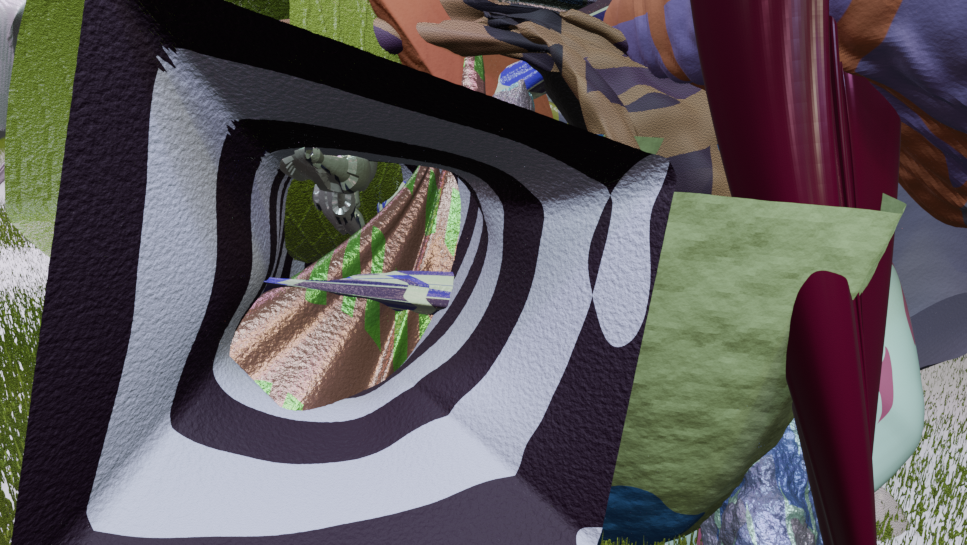} &
        \includegraphics[width=0.22\linewidth, valign=m]{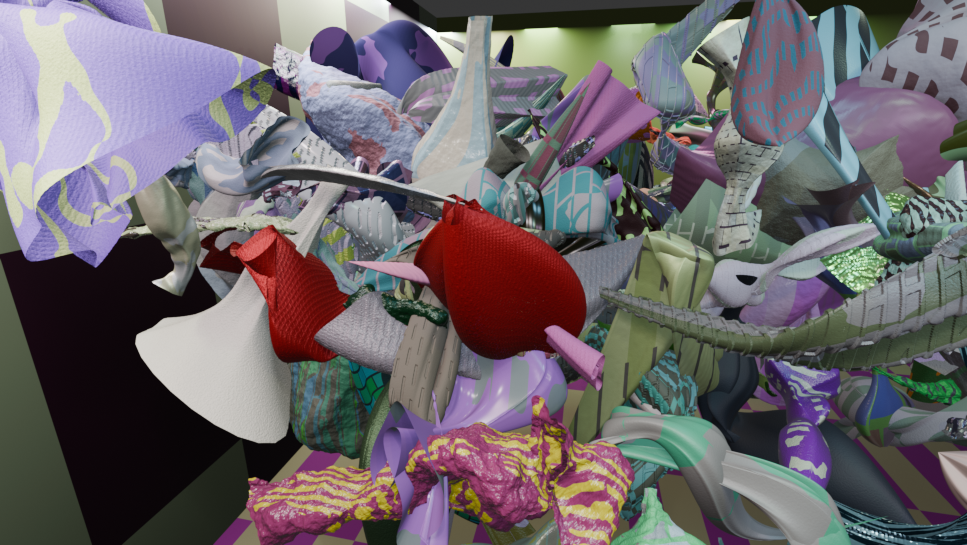} &
        \includegraphics[width=0.22\linewidth, valign=m]{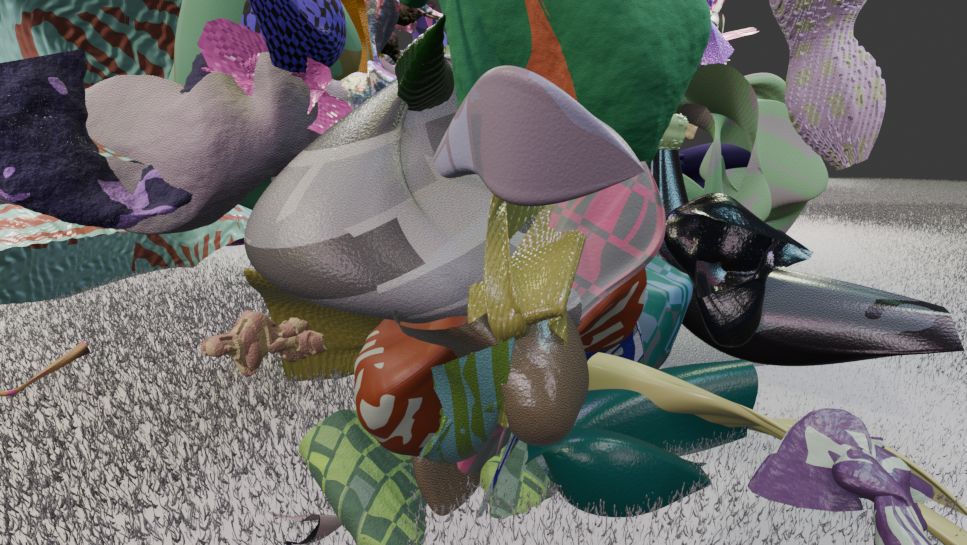} &
        \includegraphics[width=0.22\linewidth, valign=m]{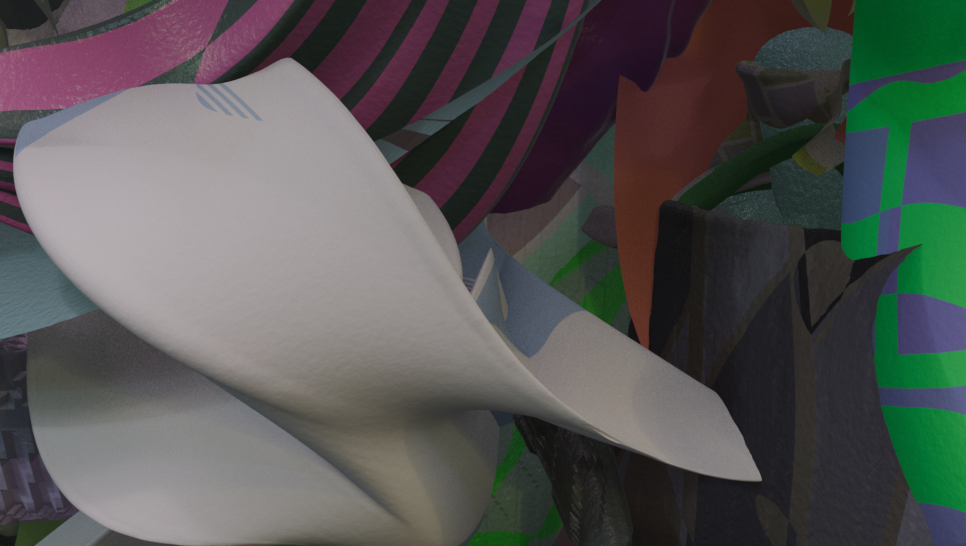} \\
        
        
        \rotatebox[origin=c]{90}{\textbf{Disparity}} & 
        \includegraphics[width=0.22\linewidth, valign=m]{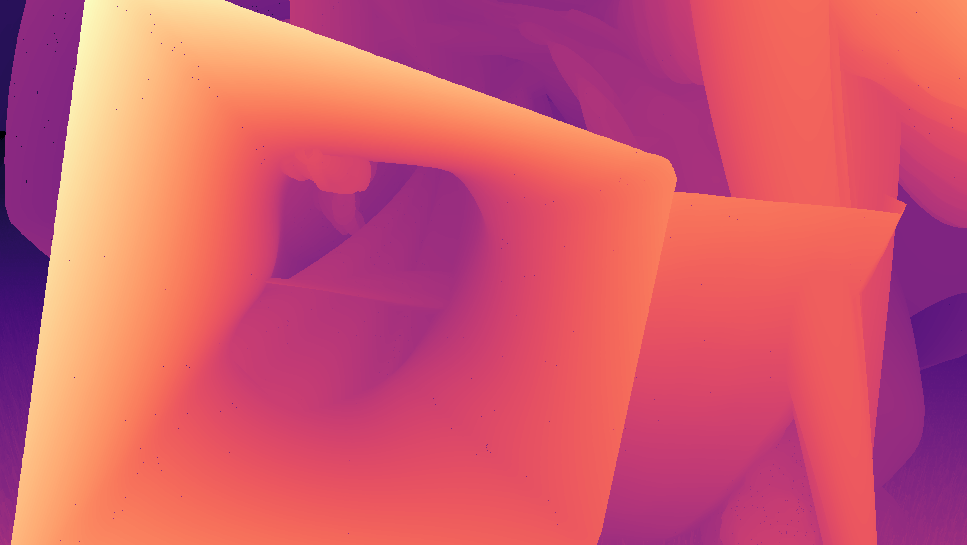} &
        \includegraphics[width=0.22\linewidth, valign=m]{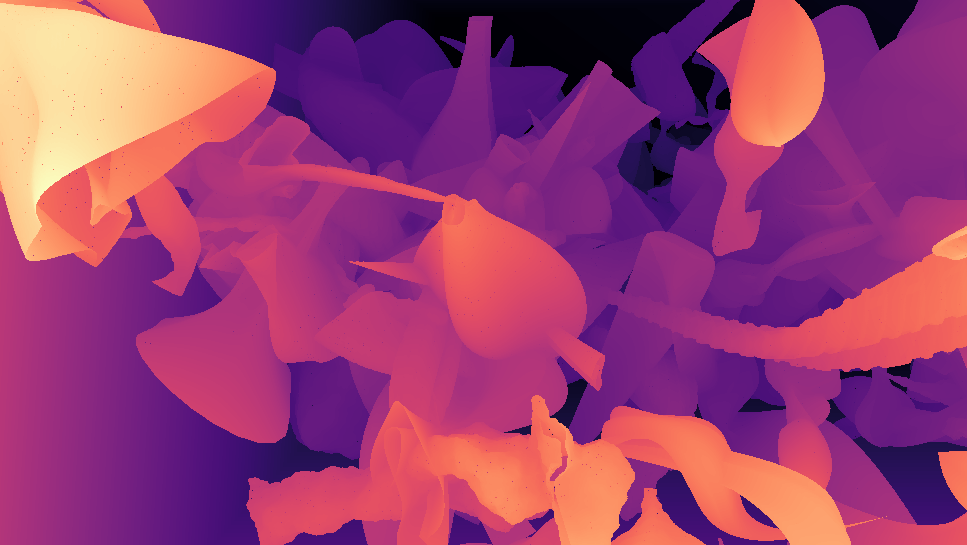} &
        \includegraphics[width=0.22\linewidth, valign=m]{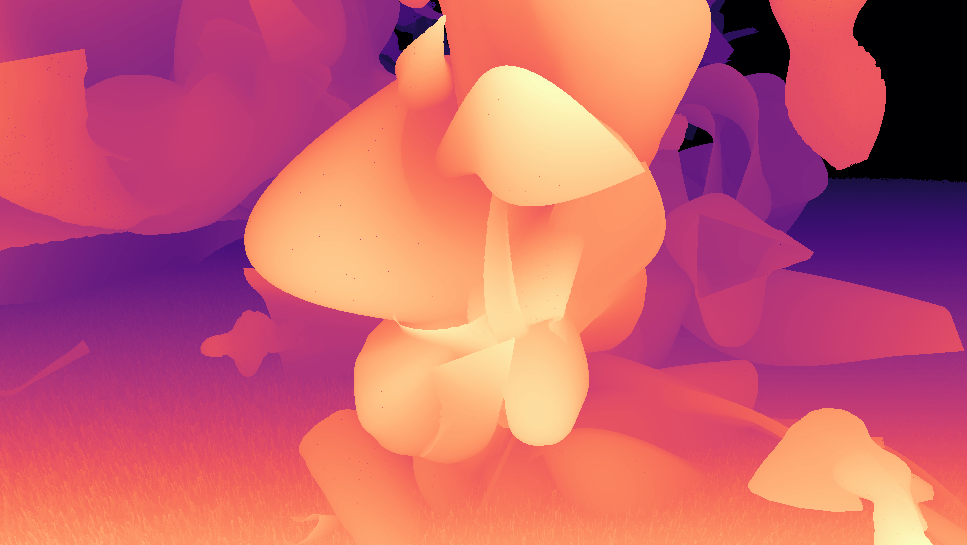} &
        \includegraphics[width=0.22\linewidth, valign=m]{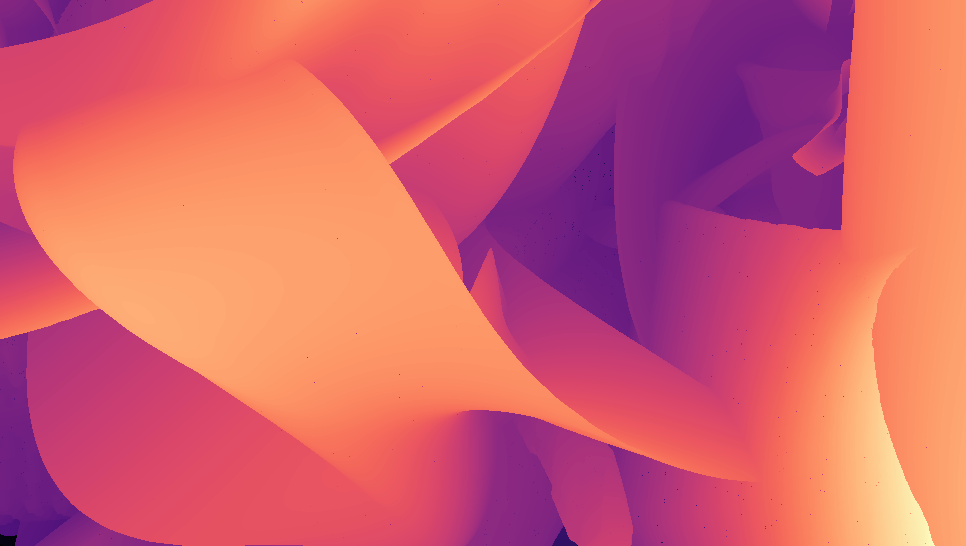} \\

    \end{tabular}
    
    \caption{Visualizing multiple scenes from $966 \times 546$ SimpleProc evaluation dataset.}
    \label{fig:simple_proc_supp}
\end{figure}
Following modifications are required from the original SimpleProc data generation to reproduce the evaluation dataset:
\begin{itemize}
  \item Modify \textit{fov\_y} and \textit{total\_pixels} parameters according to the desired width and height of the render. Remove any rounding off in image resolution.
  \item Place an anchor rig (that forms the right camera of the stereo pair) using the default pipeline and parameter sampling. 
  \item For the left rectified camera rig, we compute it's location and rotation using the anchor rig and uniformly sampled baseline shift (in range 5 to 45 cm) as:
  \begin{algorithmic}[]
   \State $right\_vec \gets anchor\_rig.matrix\_world_{3 \times 3} \cdot [1, 0, 0]^T$
    \State $cam\_location \gets anchor\_rig.location - (baseline \cdot normalized\_right\_vec)$
    \State $rig.rotation\_euler \gets anchor\_rig.rotation\_euler$
\end{algorithmic}
  \item For both camera rigs, the parent inverse of the camera object is set to identity so local camera coordinates map $1:1$ to the rig.
  \item Each scene render is performed with a unique seed value. The list of seed values used for each resolution is attached in the supplementary material. 
\end{itemize}
Example images of this generated dataset can be found in Figure~\ref{fig:simple_proc_supp}.

\subsection{Industrial Parts Evaluation Dataset}

\begin{figure*}[h]
  \centering
  \includegraphics[width=\linewidth]{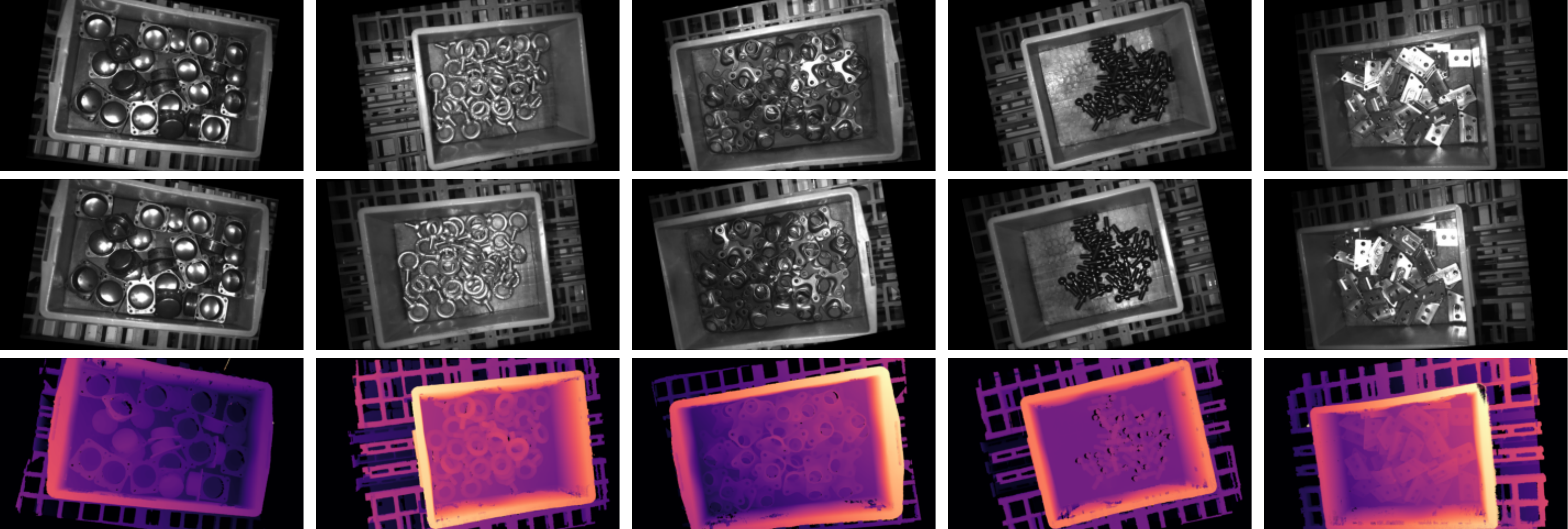}
  \caption{Examples from XYZ-IBD~\cite{huang2025xyzibdhighprecisionbinpickingdataset} dataset. Left images are the first row, followed by right images as the second row and ground truth disparity as the third row.}
  \label{fig:industrial_parts}
\end{figure*}
We include the industrial parts evaluation dataset, which is XYZ-IBD~\cite{huang2025xyzibdhighprecisionbinpickingdataset} postprocessed for stereo, as shown in Figure~\ref{fig:industrial_parts}. We include a  \textit{xyz.zip} file containing 55 folders in the supplement. In each folder there is a \textit{left\_image.jpeg}, \textit{right\_image.jpeg}, \textit{disp.png}, and \textit{calib.json}. The left and right image are already rectified and resized to 1080p. The pseudo disparity is saved in uint16 png. To get the true disparity, divide all values by 2. The calibration contains the output of cv2.stereoRectify and the baseline,  which can be used to construct a 3D pointcloud. \\ Although the ground truth for these scenes occasionally exhibits artifacts from the structured light cameras employed, we find this negligible, since these inconsistencies impact all evaluated methods in the same manner.

\begin{figure}[h]
  \centering
  \includegraphics[width=0.8\linewidth]{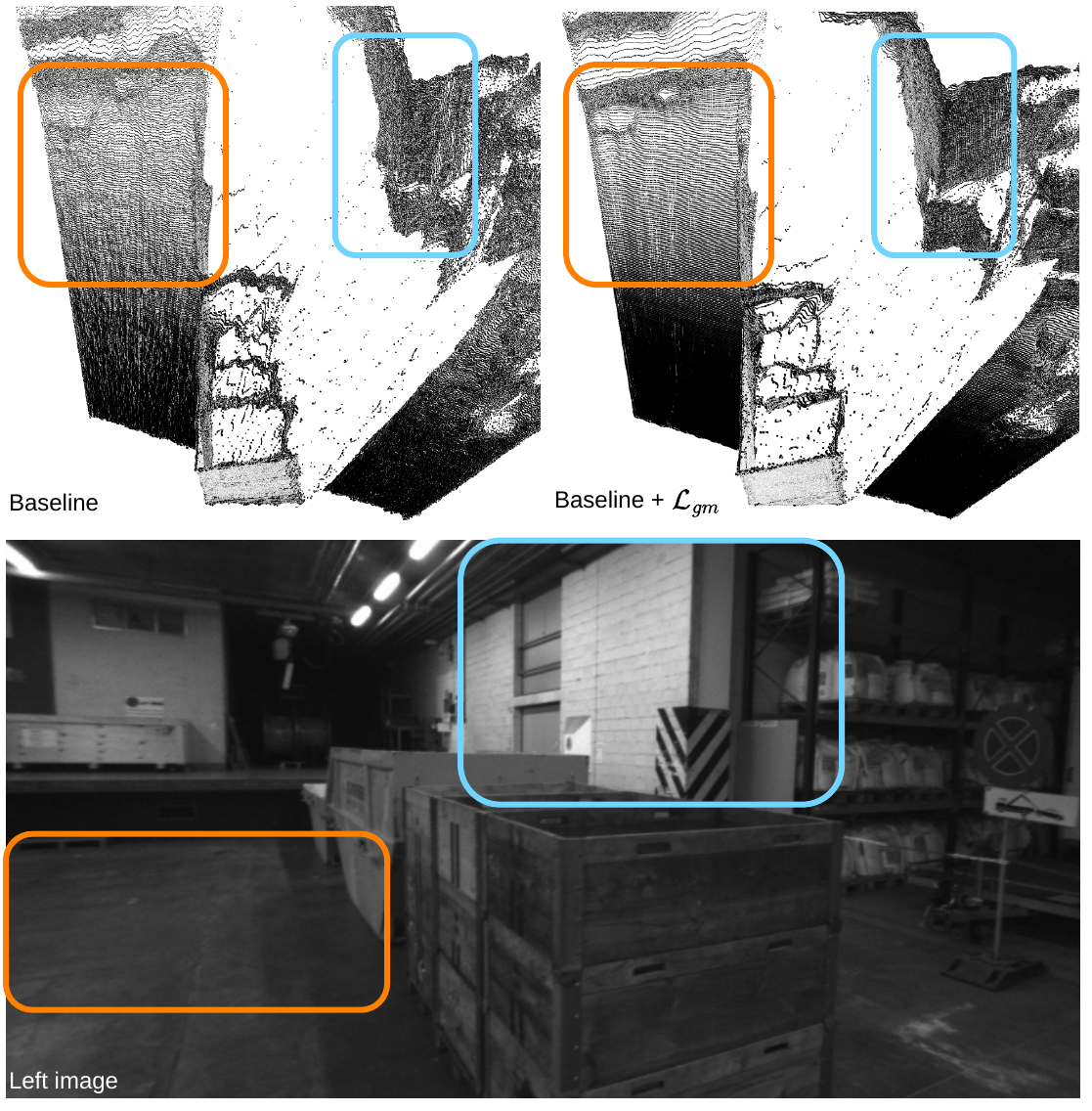}
  \caption{Qualitative comparison of disparity predictions with and without the Gradient Matching loss. The point clouds are visualized in the first row and the left image is provided in the second row. }
  \label{fig:ablation_gm_loss}
\end{figure}

\begin{table}[t]
  \caption{Ablation study on the Gradient Matching loss ($\mathcal{L}_{gm}$). While global metrics remain stable, $\mathcal{L}_{gm}$ significantly enhances local structural sharpness.}
  \label{tab:ablation_gm_loss}
  \centering
  \begin{tabular}{l ccc}
    \toprule
    Method & EPE $\downarrow$ & RMS $\downarrow$ & Bad@0.5 $\downarrow$ \\
    \midrule
    ViT-Base (Baseline) & \textbf{0.282} & 0.545 & 13.56 \\
    ViT-Base + $\mathcal{L}_{gm}$ & 0.285 & \textbf{0.525} & \textbf{13.51} \\
    \bottomrule
  \end{tabular}
\end{table}

\subsection{Ablation on Gradient Matching Loss}
Due to the significant computational requirements of training our full-scale model, this experiment was performed using a reduced parameter set (ViT-Base) and a data subset. To validate the effectiveness of the gradient matching loss, we trained this model exclusively on the FSD dataset and evaluated it on the ETH3D training split. Table~\ref{tab:ablation_gm_loss} and Figure~\ref{fig:ablation_gm_loss} summarize the ablation results. Although the impact on global metrics is minor, the qualitative results in Figure~\ref{fig:ablation_gm_loss} demonstrate that $\mathcal{L}_{gm}$ is essential for suppressing noise on the geometry reconstructions. With $\mathcal{L}_{gm}$, the ground and the walls are less noisy. By enforcing sharp depth discontinuities, the loss prevents the structural blurring seen in the baseline, which is critical for high-fidelity 3D point cloud generation.

\subsection{ETH3D Leaderboard}

\begin{figure*}[h]
  \centering
  \includegraphics[width=\linewidth]{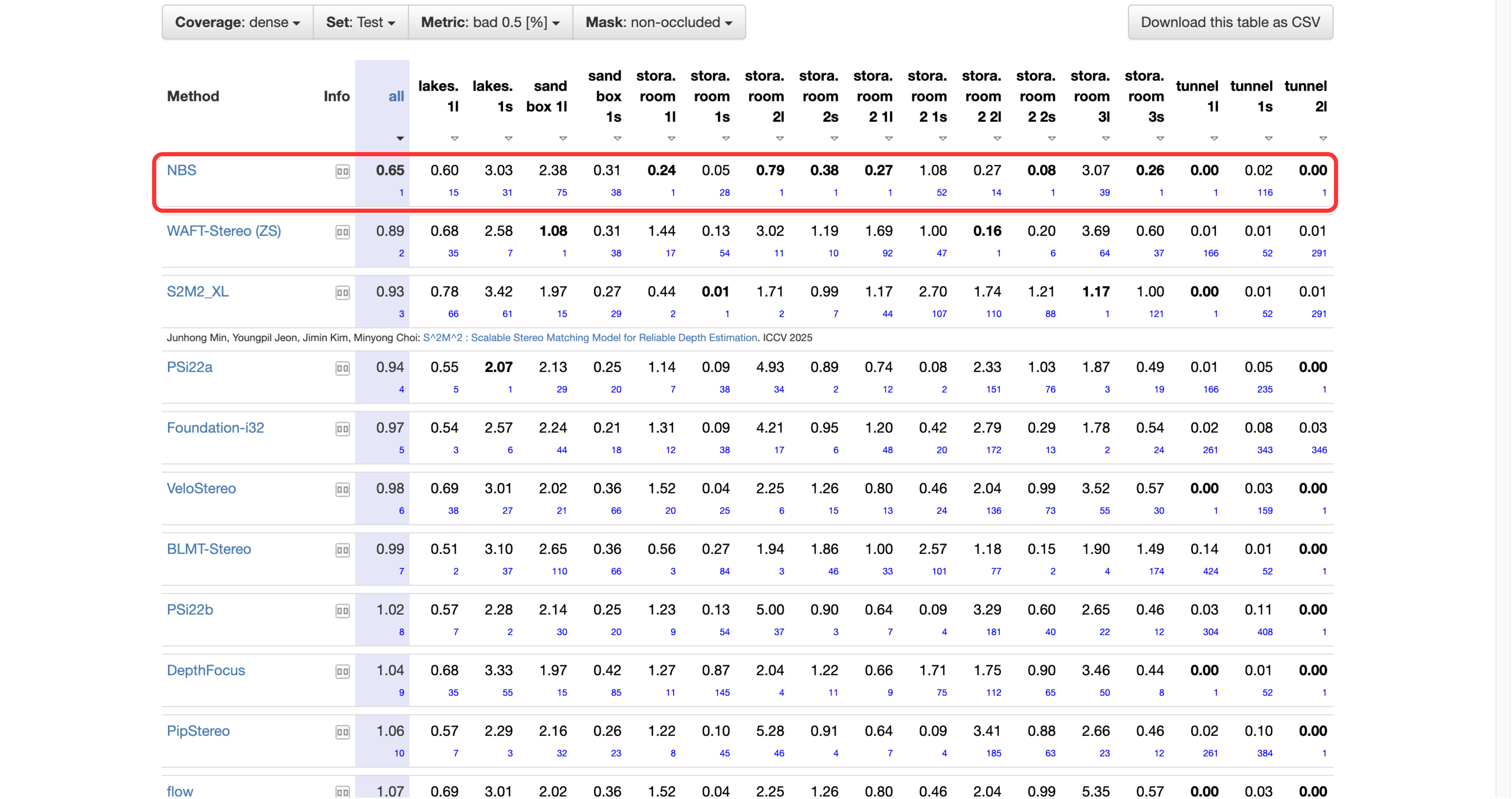}
  \caption{ETH3D leaderboard~\cite{eth3d_stereo} screenshot of bad@0.5 for non occluded pixels. Our model ranks first.}
  \label{fig:eth3d_0_5_non_occ}
\end{figure*}

\begin{figure*}[h]
  \centering
  \includegraphics[width=\linewidth]{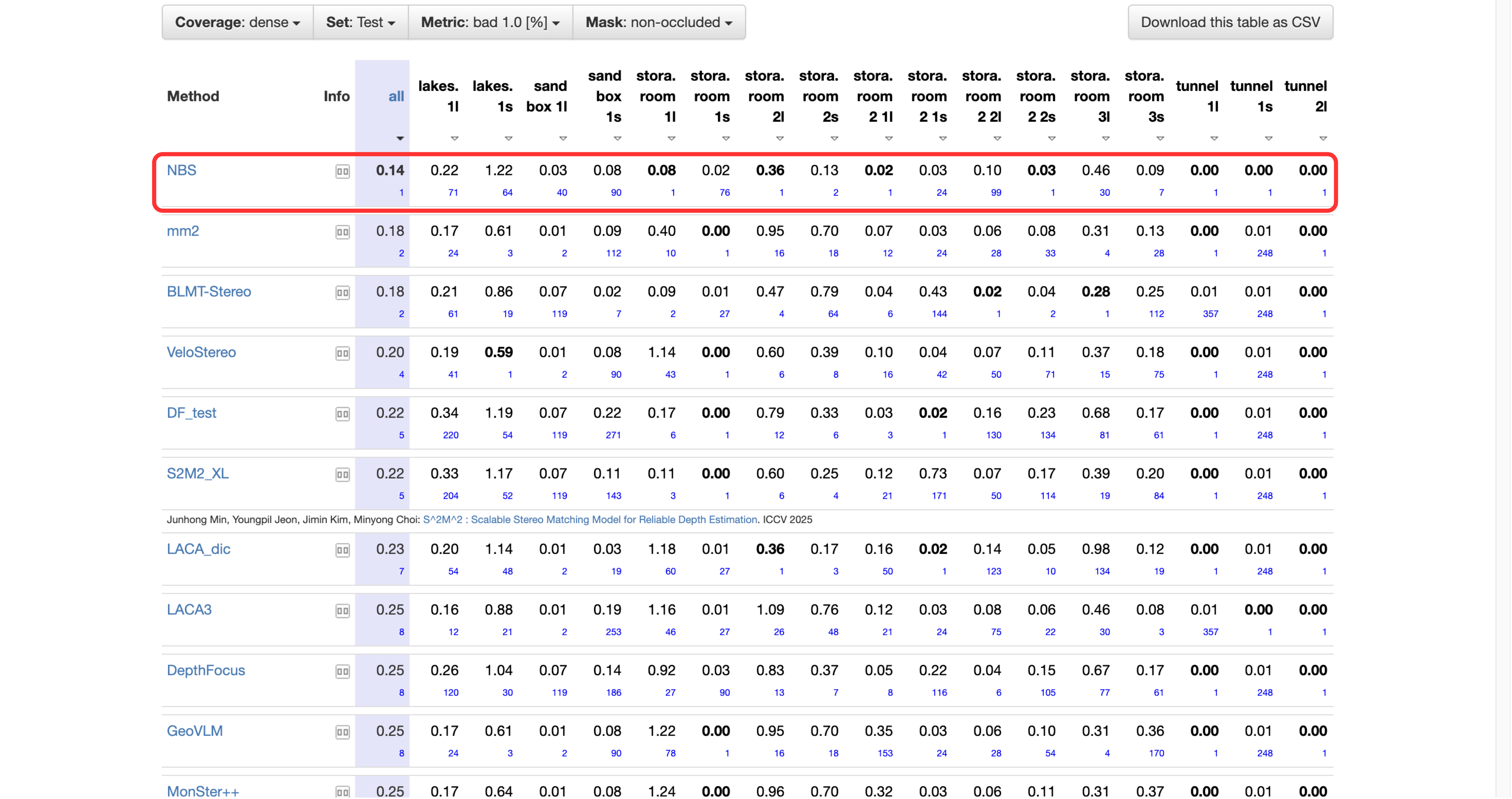}
  \caption{ETH3D leaderboard~\cite{eth3d_stereo} screenshot of bad@1 for non occluded pixels. Our model ranks first.}
  \label{fig:eth3d_1_non_occ}
\end{figure*}

\begin{figure*}[h]
  \centering
  \includegraphics[width=\linewidth]{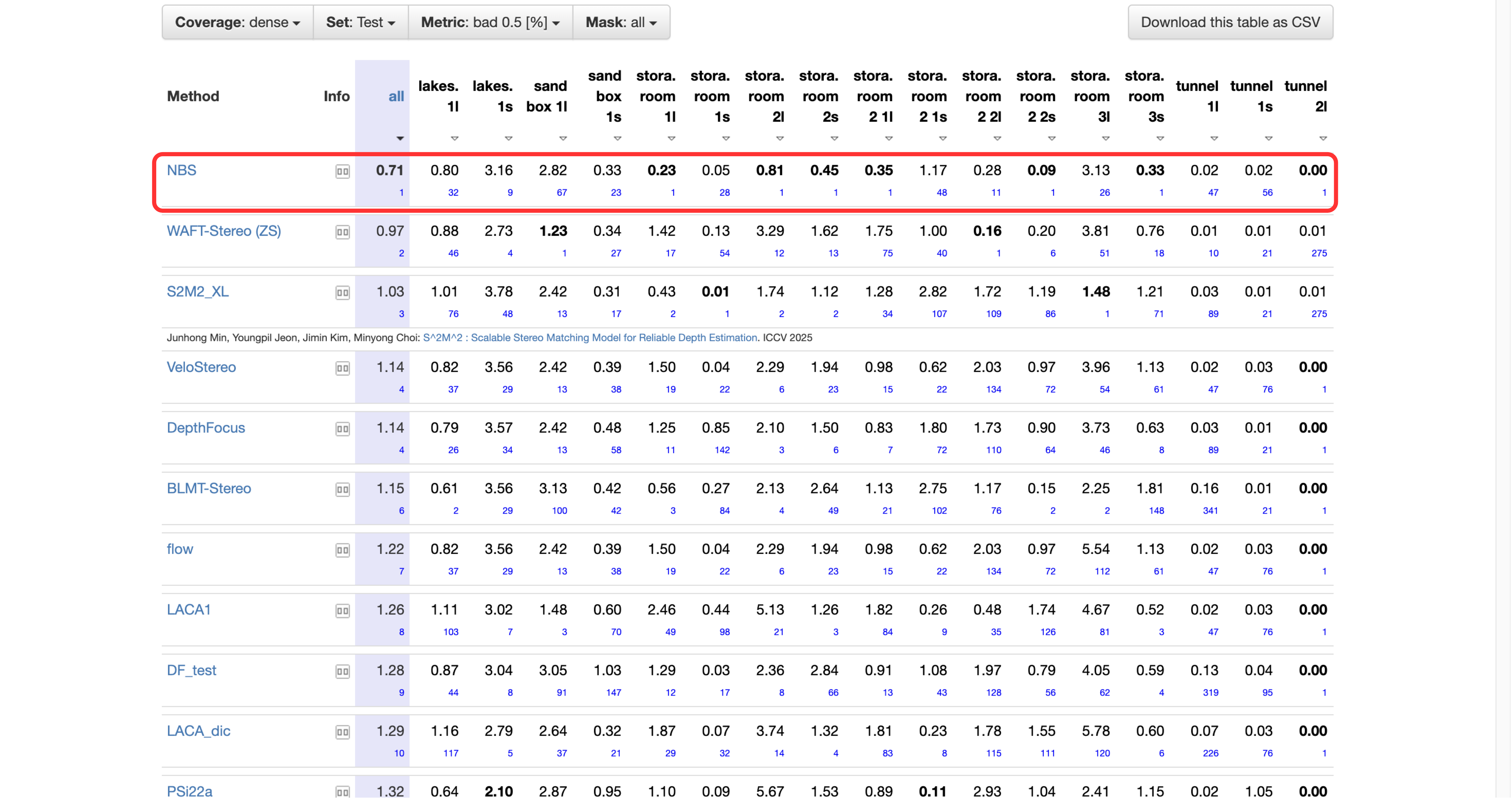}
  \caption{ETH3D leaderboard~\cite{eth3d_stereo} screenshot of bad@0.5 for all pixels. Our model ranks first.}
  \label{fig:eth3d_0_5_all}
\end{figure*}

\begin{figure*}[h]
  \centering
  \includegraphics[width=\linewidth]{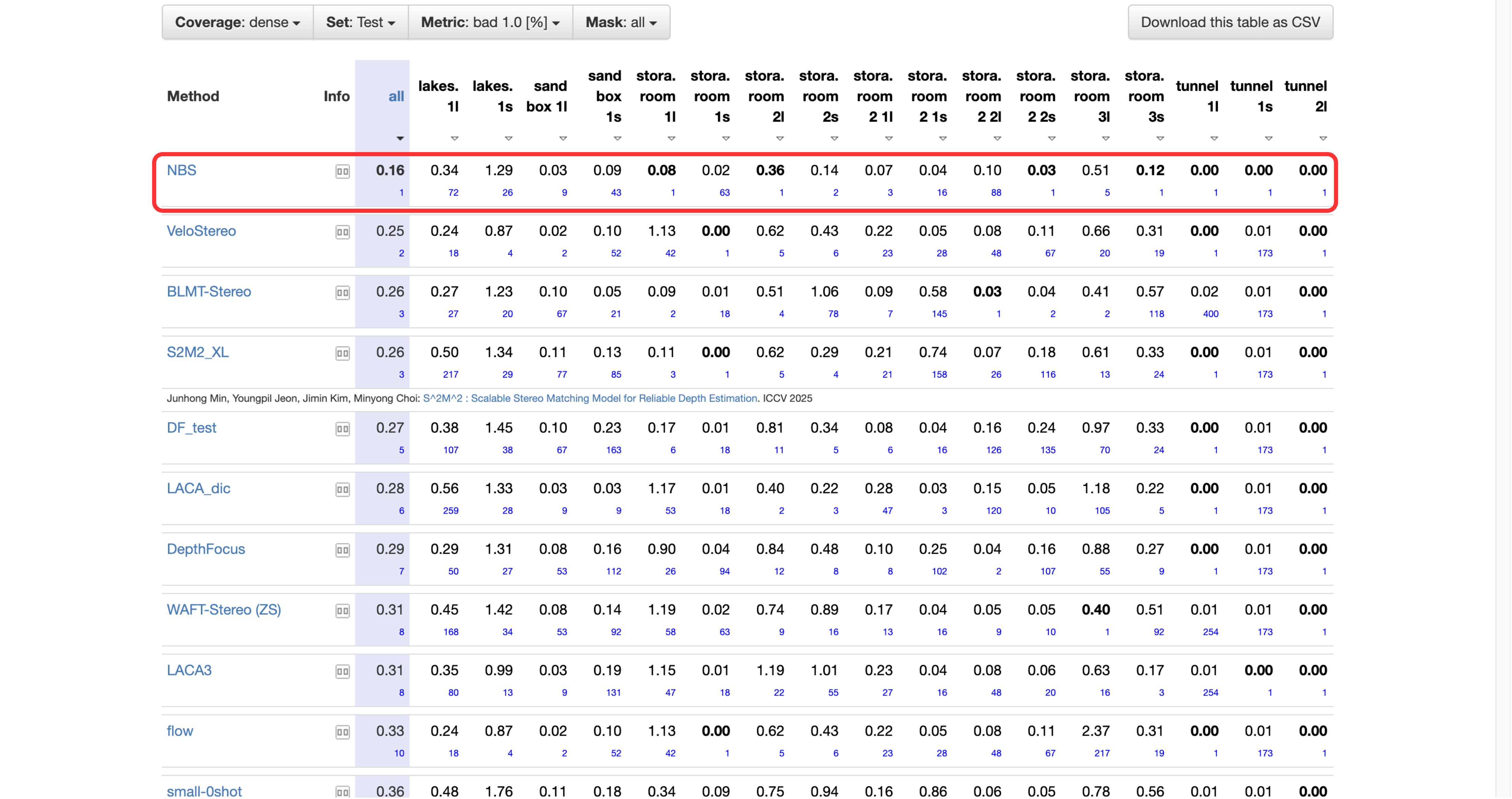}
  \caption{ETH3D leaderboard~\cite{eth3d_stereo} screenshot of bad@1 for all pixels. Our model ranks first.}
  \label{fig:eth3d_1_all}
\end{figure*}

At the time of submission, our method achieved first place on the ETH3D Leaderboard~\cite{eth3d_stereo}. Figures~\ref{fig:eth3d_0_5_non_occ},~\ref{fig:eth3d_1_non_occ},~\ref{fig:eth3d_0_5_all}, and~\ref{fig:eth3d_1_all} demonstrate that our model consistently outperforms existing approaches across multiple metrics by a substantial margin. We provide screenshots of only a subset of the metrics, while Table~\ref{tab:eth3d-results_supp} shows the results for all metrics.

\end{document}